\documentclass[letterpaper, 10 pt, conference]{ieeeconf}

\IEEEoverridecommandlockouts
\usepackage{xcolor}
\usepackage{times}
\usepackage{epsfig}
\usepackage{graphicx}
\usepackage{tabularx,ragged2e,booktabs}

\usepackage{amsmath}
\usepackage{amssymb}
\usepackage{bbm}
\usepackage[misc]{ifsym}
\usepackage[export]{adjustbox}
\usepackage{float}
\usepackage{wrapfig}
\usepackage{lscape}
\usepackage{rotating}
\usepackage{epstopdf}

\usepackage{verbatim}
\usepackage{booktabs}
\usepackage{multirow}
\usepackage{makecell}
\usepackage{authblk}
\usepackage{hyphenat} 

\usepackage{pgf}

\usepackage{pifont}
\usepackage{times}
\usepackage{soul}
\usepackage{url}
\usepackage{graphicx}
\usepackage{amsmath}
\usepackage{booktabs}
\usepackage{balance} 
\usepackage[T1]{fontenc}    
\usepackage{wrapfig}
\usepackage{fancybox}
\usepackage{algpseudocode}
\usepackage{algorithmicx}
\usepackage[ruled,vlined,linesnumbered,noend]{algorithm2e}
\usepackage{diagbox}
\usepackage{multirow}
\usepackage{multicol}
\usepackage{amsfonts}       
\usepackage{nicefrac}       

\usepackage{graphbox}
\usepackage{subcaption}
\usepackage{multicol}

\usepackage[colorlinks=true]{hyperref}

\graphicspath{{figures/}}

\begin{document}
\title{Object-Oriented Material Classification and 3D Clustering for Improved Semantic Perception and Mapping in Mobile Robots}

\author{Siva Krishna Ravipati$^1$  \and Ehsan Latif$^2$ \and Ramviyas Parasuraman$^{1,2,*}$ \and Suchendra M. Bhandarkar$^{1,2}$
\thanks{$^1$ Institute for Artificial Intelligence, University of Georgia, Athens, GA 30602, USA}
\thanks{$^2$ School of Computing, University of Georgia, Athens, GA 30602, USA }
\thanks{$^*$ Corresponding author email: \tt{ramviyas@uga.edu}}
}

\maketitle              

\begin{abstract}
Classification of different object surface material types can play a significant role in the decision-making algorithms for mobile robots and autonomous vehicles.
RGB-based scene-level semantic segmentation has been well-addressed in the literature. However, improving material recognition using the depth modality and its integration with SLAM algorithms for 3D semantic mapping could unlock new potential benefits in the robotics perception pipeline.
To this end, we propose a complementarity-aware deep learning approach for RGB-D-based material classification built on top of an object-oriented pipeline. The approach further integrates the ORB-SLAM2 method for 3D scene mapping with multiscale clustering of the detected material semantics in the point cloud map generated by the visual SLAM algorithm. Extensive experimental results with existing public datasets and newly contributed real-world robot datasets demonstrate a significant improvement in material classification and 3D clustering accuracy compared to state-of-the-art approaches for 3D semantic scene mapping.

\end{abstract}

\IEEEpeerreviewmaketitle

\section{Introduction}
\label{sec:intro}


Recent advancements in mobile robotics have underscored the importance of autonomous navigation and manipulation in unknown environments. A pivotal challenge in this domain is the accurate identification and classification of surface materials of objects, a capability crucial for effective decision-making and interaction within these environments, whether for exploration, manipulation, or clearing tasks \cite{andreas2015multimodaldl}.
Specifically, the deployment of robots in exploration and mapping applications \cite{latif2023seal,kannan2020material} benefits from accurate perception and understanding of the environment, especially when integrated with SLAM (Simultaneous Localization and Mapping) algorithms \cite{murORB2}.

Whether in domestic settings, firefighting, or logistics, understanding the material composition of surroundings is crucial for preplanning operations and navigating effectively \cite{alexnet,kannan2020material}. 
For instance, distinguishing between concrete and black ice is essential for safely operating self-driving vehicles and service robots. 
Incorporating material recognition into conventional object recognition, scene understanding, and SLAM pipeline can significantly enhance robot performance, especially in use cases involving physical interactions and realistic renderings in virtual environments \cite{semfusion2017map,schwartz2013pixelMat}. 

\begin{figure}[t]
\centering
 \includegraphics[width=0.72\linewidth]{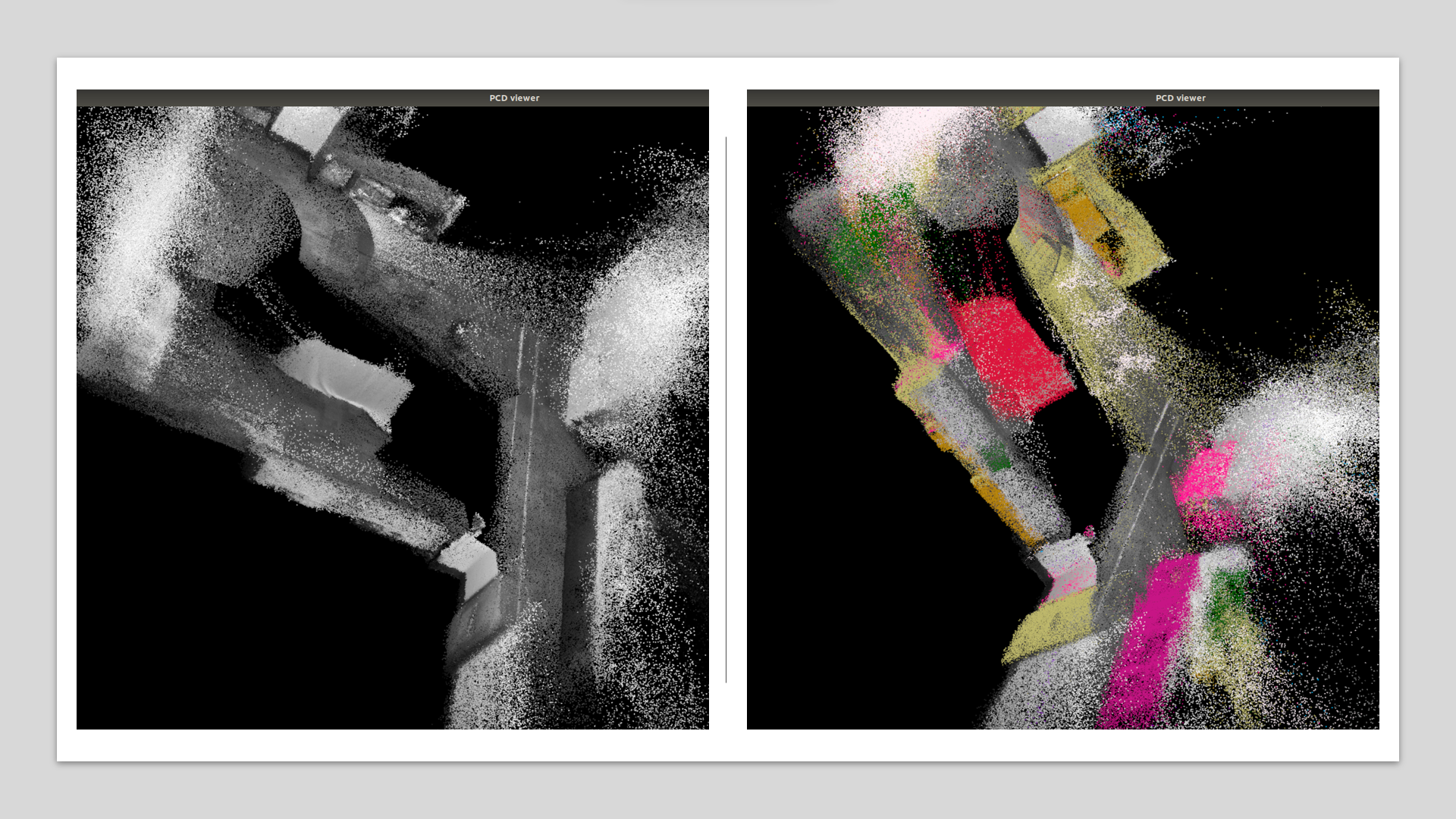}
  \includegraphics[width=0.25\linewidth]{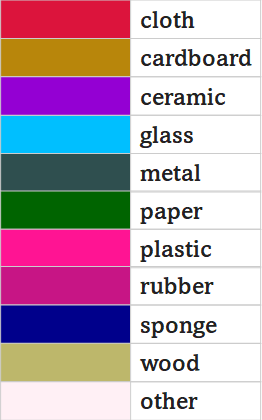}
 \caption{Illustration of the objected-oriented 3D semantic mapping with material-level information  (right) based on the RGB-D point clouds (left), shown along with the labels.}
 \label{fig:semantic_map}
 \vspace{-4mm}
\end{figure}

Current computer vision methods for material classification often focus on visual cues (shape and color) from RGB images.
However, traditional RGB-based scene-level semantic (material) segmentation methods provide limited insights into the surface properties of the objects. While generally effective, they fall short in providing the nuanced material recognition necessary for robotic tasks, thus motivating the use of depth modality (i.e., RGB+D) for extracting rich features \cite{Gupta2014LearningRF}.
The challenge lies in effectively utilizing depth maps to predict material types from camera images, which becomes critical in complex and dynamic environments \cite{Lee2019IRreflectance}.
Therefore, creating a semantic material map using RGB-D images presents a valuable research avenue. Such maps enhance scene understanding and aid in robotic exploration and manipulation.
Additionally, point cloud mapping, which creates a 3D representation of the environment, is invaluable for mapping and navigation, as well as in generating detailed models for mixed reality and architectural planning \cite{murORB2}.

Motivated by the need to enhance robotic perception \cite{yang2021can}, particularly in the context of material classification, we introduce a novel approach to material classification and semantic mapping for mobile robots. Our framework presents a unique result, as illustrated in Fig.~\ref{fig:semantic_map}, which showcases the point cloud output combined with object-oriented material identification and clustering.
We integrate our approach into the well-recognized ORB-SLAM2 algorithm \cite{murORB2} leveraging the benefits of visual odometry and SLAM, providing a real-time, accurate global map essential for mobile robots. 

The key contributions in this paper include:
\begin{itemize}
    \item A novel material classification network through complementarity-aware fusion of RGB and Depth-based convolutional neural networks, built on top of an RGB-based object detection pipeline for fast (real-time) and accurate material classification. The approach takes advantage of the extraction and fusion of distinctive and correlated features from RGB and depth modalities.
    \item Integration of the material classification outcome with the RGB-D SLAM using a voxel-based multiscale feature matching technique to obtain a precise metric-semantic mapping, a 3D environmental map consistently clustered by the material properties of the objects.
    \item An extensive experimental evaluation of our architecture with multiple real-world datasets (standard and custom) on material classification and 3D semantic mapping of complex environments showing up to 15\% improved accuracy over state-of-the-art (SOTA) approaches.
\end{itemize}

Finally, we contribute new real-world mobile robot RGB-D datasets with meaningful object and material classes (as ROS bags) and open source the relevant codes\footnote{\url{https://github.com/herolab-uga/matsee}} to benefit the community. 
Equipping robots with the ability to discern and cluster material properties accurately enables more reliable and efficient task execution across a range of applications, from domestic service to industrial automation \cite{Degol_2016}. 

\section{Related Work}
\label{sec:relatedwork}
Recent advancements in vision-based SLAM and material recognition have leveraged deep learning to achieve notable progress. 
The complexity of indoor environments and the diverse material composition of objects therein make this an especially challenging and relevant problem \cite{Zhao2017AFE, Lee2019IRreflectance,latif2022multi}. 
Mur-Artal and Tardós \cite{murORB2} introduced ORB-SLAM2, an efficient SLAM pipeline for monocular, stereo, and RGB-D cameras. It provides high accuracy but lacks semantic understanding. In the domain of material recognition, Qi et al. \cite{qi2017pointnet2} proposed PointNet++, which directly processes point clouds for 3D classification and segmentation. Although they process point clouds efficiently, these models do not consider the material properties of objects. Our approach complements this by providing material-level semantic information.

Chen et al. \cite{chen2018CAFusion} developed a progressively complementarity-aware fusion network for RGB-D salient object detection. Their method effectively fuses RGB and depth features but is not tailored for material classification. Our proposed CA fusion module extends this idea to material recognition, enhancing the accuracy of object-material association. Schwartz and Nishino \cite{schwartz2016matRecog} focused on material recognition from a global context, emphasizing the role of local features. Their work, while insightful, does not incorporate depth information, which is critical for distinguishing materials with similar textures. By incorporating depth data, our method provides a more robust solution for material classification in complex scenes.

In robot mapping, Zhao et al. \cite{Zhao2017AFE} implemented a deep learning method for 3D reconstruction and material recognition, yet their approach does not effectively handle dynamic environments. Similarly, the study in \cite{Zhao2020SimultaneousMS} achieves simultaneous material segmentation and 3D reconstruction, primarily in static industrial settings, but it falls short in adapting to changing environments that are integral to mobile robots. Our method addresses these limitations with a voxel-based matching component, significantly enhancing the SLAM system's adaptability in dynamic scenarios. This is a critical improvement over existing methods, as it enables accurate real-time mapping and material recognition in environments where conditions and object placements are constantly evolving. Furthermore, works in \cite{semfusion2017map,hempel2022online,Martins2020ExtendingMW} contributed to integrating object-level semantic information with SLAM to improve localization accuracy through graph-based pose corrections, but these methods do not differentiate different materials. While we do not focus on improving the localization, our approach fills the gap by adding material-level semantic maps with a vision to extend the capabilities of robotic systems in complex, variable settings by providing a detailed and dynamic semantic material map essential for interaction tasks such as grasping and manipulation.


In a recent work \cite{OnlineObjSemMap}, the authors proposed an online 2D and 3D semantic, modular map representation and object detection framework using RGB-D data over-refinement and likelihood maintenance to avoid false detection. However, the object-based likelihood maintenance mechanism may miss out on semantically important objects that are occluded by other objects and have a low hit-to-miss ratio in likelihood calculation. In our approach, we use a voxel-based feature matching technique, which considers all the objects irrespective of their occurrence frequency.

In summary, our approach departs from the above SOTA methods by integrating RGB-D data fusion and clustering object-level results with point clouds for consistent semantic mapping in 3D environments. Leveraging a complementarity-aware (CA) fusion module, our system synergistically combines RGB and depth data, enabling more refined material classification than traditional models that treat these modalities in isolation. This method captures unique visual characteristics and incorporates depth information, crucial for discerning materials with similar appearances under varying lighting conditions. Unlike existing methods that might overlook material attributes in environmental mapping, our approach recognizes and classifies materials with improved accuracy.
Additionally, using RGB-D images can improve the system's robustness to lighting changes and provide more accurate representations of the environment in low-light conditions.

Another key aspect of our method is incorporating voxel-based matching within the SLAM framework. This component ensures both dynamic and static objects in the point cloud map are correctly identified and associated with their material types. Such an advancement is vital in robotics, where accurate material recognition influences tasks like manipulation or navigation. Our approach extends the capabilities of existing systems like ORB-SLAM2 \cite{murORB2}, PointNet \cite{qi2017pointnet2}, and SemanticFusion \cite{semfusion2017map} by addressing their limitations in depth and semantic understanding and maintaining the consistency of semantic mapping in 3D, a feature not fully realized in prior works. Consequently, our approach offers a more robust, real-time semantic material map, greatly enhancing a robot's perception and interaction abilities in complex and dynamically changing environments.
As a result, combining object detection and point cloud mapping with RGB-D images can provide a rich and detailed representation of the environment, enhancing performances in various tasks such as localization, navigation, exploration, and object manipulation \cite{latif2023seal,Schwarz_2017}. 
In addition to these advantages, using RGB-D images to create a semantic material map also has several other benefits. For example, the inclusion of depth information in the RGB-D images can improve the accuracy of material identification and localization and can also provide additional information about the shape and texture of objects in the scene. Additionally, using RGB-D images can improve the system's robustness to lighting changes and provide more accurate representations of the environment in low-light conditions. These benefits can also be extended to a multi-robot system, providing with enhanced localization and navigation capabilities \cite{latif2022dgorl,latif2022multi}.

\begin{figure*}[t]
    \centering
    \includegraphics[width=\linewidth]{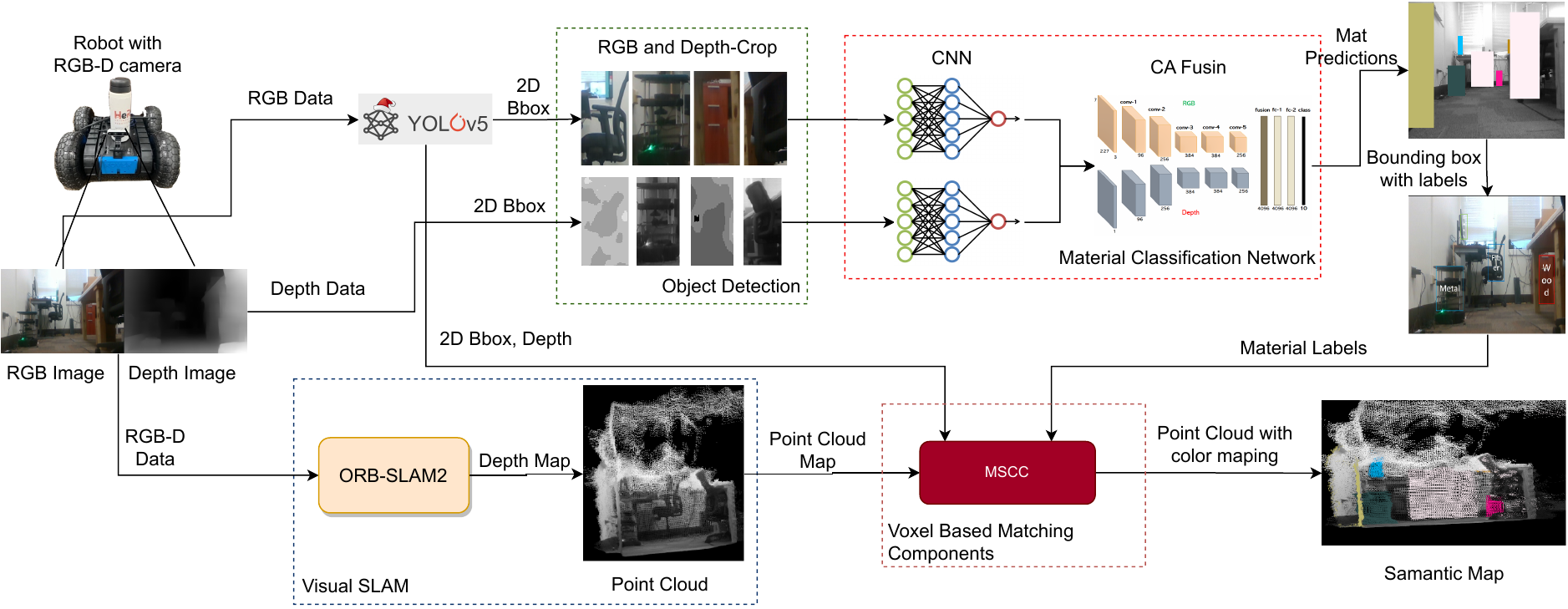}
    \caption{Architectural overview of the proposed object-oriented 3D semantic mapping with material labels. The visual SLAM ($\mathtt{VSLAM}$) component generates the point cloud map, and the YOLO component ($\mathtt{OBJ}$) detects objects and locates the bounding boxes of the objects in the images. The material classification network ($\mathtt{MCN}$) classifies the objects in the bounding boxes into different material classes. The voxel-based matching component ($\mathtt{VOXM}$) uses the point cloud map generated by visual SLAM and the material labels obtained from the material classification component to match the 3D coordinates of the bounding boxes with the 3D coordinates of the point cloud and propagate the material labels to the points in the point cloud.}
    \label{fig:architecture}
    \vspace{-4mm}
\end{figure*}

\section{Proposed Approach}
\label{sec:problem}

The core problem we address is the semantic mapping of materials in a 3D environment captured by a mobile robot. Mathematically, this involves identifying and classifying various materials within the robot's field of view, represented by a 3D point cloud. Let $\mathcal{P}$ denote the point cloud, where each point $p_i \in \mathcal{P}$ has associated $f_{RGB}$ color and $f_{Depth}$ depth feature map obtained through fusion mechanisms. The goal is to assign a material label $l_i \in \mathcal{L}$ to each point $p_i$ associated with fused map $f_{fuse}$, forming clusters of points (map) with similar material properties.

The proposed solution involves three main pillars. It first detects objects in the environment using a fast object detection pipeline (Sec.~\ref{sec:yolo}), followed by material classification of these detected objects (Sec.~\ref{sec:material}). Using these classified features with a 3D point cloud map generated by a visual SLAM (VSLAM) algorithm (e.g., ORB-SLAM2), a feature matching technique described in Sec.~\ref{sec:mscc} is applied to create a comprehensive semantic map of the environment. An architectural overview of the proposed approach is shown in Fig.~\ref{fig:architecture}. 
This integrated approach is designed to enhance the perception capabilities of mobile robots by enabling them to recognize and accurately classify materials in their environment. 
Algorithm~\ref{alg:semantic-clustering} provides the pseudocode description of our semantic mapping of materials from the RGB-D data.

\begin{algorithm}[tb]
 \SetAlgoLined
 Input: RGB-D data stream\;
 Models: Object detection model ($\mathtt{OBJ}$), material classification network ($\mathtt{MCN}$), visual SLAM framework ($\mathtt{VSLAM}$), and voxel-based matching component ($\mathtt{VOXM}$)\;
 Return: 3D semantic map $\mathcal{SM}$ (point cloud clusters) with object and material labels\;   
 \While{robot is navigating the environment}{
    Capture current RGB frame and depth data\;
    Detect objects using $\mathtt{OBJ}$ on the RGB frame and generate 2D bounding boxes\;
    \For{each detected object}{
        Crop corresponding RGB and Depth sections\;
        Classify the material using $\mathtt{MCN}$\;
        Assign material label $l_i$ to the object\;
    }
    Generate a sparse 3D point cloud map ($\mathcal{P}$) of the environment using $\mathtt{VSLAM}$\;
    Divide the point cloud $\mathcal{P}$ into a voxel grid $\mathcal{V}$\;
    \For{each voxel in $\mathcal{V}$}{
        Find the closest 3D bounding box from $\mathtt{OBJ}$ output\;
        Propagate the corresponding $l_i$ outputs from $\mathtt{MCN}$ to points within the voxel\;
    }
    Apply the 3D clustering algorithm ($\mathtt{VOXM}$) on $\mathcal{P}$ to obtain segmented point clouds $\mathcal{PS}$\;
    Propagate material labels $l_i$ to the clusters in $\mathcal{PS}$\;
    Update semantic map $\mathcal{SM}$ with material $l_i$ (and optionally, object labels) for each point\;
    }
 \caption{3D Semantic Mapping of Materials}
 \label{alg:semantic-clustering}
\end{algorithm}

\subsection{Object Detection ($\mathtt{OBJ}$) Pipeline}
\label{sec:yolo}
Our approach uses an object-oriented pipeline, meaning that the object detections trigger the material classifications. Input RGB images (from an RGB-D camera) are used for object detection, and we build our architecture on top of the SOTA YOLO (You Only Look Once)  model~\cite{redmon2016you} because of its fast, robust, accurate, and versatile real-time object detection capabilities. We specifically used the YOLOv5 version, as it had proven to be robust in the ROS framework\footnote{\url{https://ros.org/}} (for SLAM integrations) with excellent performance on several benchmarks, including the COCO object detection dataset \cite{lin2014microsoft}. Regardless, the research community has consistently upgraded the YOLO-based models. For instance, the newer YOLOv8 \cite{yolov8} can replace the YOLOv5 in our pipeline. In our approach, we have utilized feature pyramid networks, which allow the model to process inputs at numerous scales and produce multiscale predictions. This process enables the model to detect small and large objects in the same image, which can be challenging for other object detection models. The output of YOLO (bounding boxes of all detected objects along with their labels) is used further for material classification and localization in a 3D semantic map. We have customized the YOLO model (building on top of a pre-trained model with the COCO dataset \cite{lin2014microsoft}) to add five new object classes: \textit{board, door, mat, robot, and trash bin}, as these new classes are repetitive in our academic office/lab settings, and will be used in the later mapping procedure.

\subsection{Material Classification Network ($\mathtt{MCN}$)}
\label{sec:material}
Once the objects are detected, each object is cropped from the RGB and the aligned Depth images using the object's bounding boxes. 
Because of the different image generation mechanisms between RGB and depth images, fusing cross-modal features effectively is a key issue for RGB-D-based material classification.
In our work, we exploit the concept of complementarity-aware (CA) fusion proposed in \cite{chen2018CAFusion} as it effectively merges the distinct features from RGB and depth data, addressing the limitations of relying solely on the visual appearance in RGB images. By processing RGB and depth images independently, our CA fusion for material classification extracts diverse features, capturing various aspects of the material types and enhancing accuracy by fusing distinctive and correlative features from the two modalities (color and depth). 
While recent works such as \cite{ding2019DepthAware} have explored the concept of \textit{late fusion} of classification outcomes from multiple modalities, they might yield some specific patterns in multiple modalities instead of finding shared common modal patterns. In a late fusion, the classifications are obtained from two parallel networks adopted to learn saliency maps from the high-level features of RGB and depth images separately. These are then concatenated to obtain a final prediction map. In contrast, the CA fusion mechanism encourages the determination of complementary information from the different modalities at different abstraction levels.

To excavate the complementarity of two modalities and maintain the discriminability of cross-modal features, we use a Complementarity-aware Fusion Network (CAFN). We first model the distinctive features from two modalities, then select complementary information of two modality features in spatial dimension with two symmetry gates. Finally, an element-wise weighting mechanism is conducted to fuse them to capture more discriminative cross-modal features. The fused features retain not only information existing in both modalities but also modality-specific information. This reduces fusion ambiguity and increases fusion efficiency. In principle, CAFN can be extended to include other modalities as well (e.g., depth-aligned LIDAR data). 

In our implementation, CAFN includes two symmetric backbones for RGB and depth feature extraction and five cascaded fusion modules. 
We use ResNet-101 as unimodal symmetric backbones similar to \cite{wu2022complementarity}. We remove the average pooling and the fully connected layers of the backbone. The last two stages are modified with dilated convolution to maintain feature resolution for more spatial information. Then we use hierarchical features from RGB and depth branches respectively,  i.e., \{$F^{i}_{RGB}$ $|$ i=1,2,3,4,5\} and \{$F^{i}_{Depth}$ $|$ i=1,2,3,4,5\} {with five cascade layers}. 
Two unimodal features $F^{i}_{RGB} \in  R^{C_i \times H_i \times W_i}$ and $F_{Depth}^i \in R^{C_i \times H_i \times W_i}$ extracting from corresponding backbones are sent to fusion modules, where $C_i$, $H_i$ and $W_i$ refer to the channel, height, and width number of the $i^{th}$ layer respectively.
As a result, CAFN can select complementary information from two modalities and then fuse enhanced unimodal features for accurate cross-modal features with the help of multiple cascaded layers. 

Let $\mathbf{I}_{RGB}$ and $\mathbf{I}_{Depth}$ be the RGB and Depth input images, respectively, and $f_{RGB}$ and $f_{Depth}$ be their corresponding CNN feature maps which can be obtained by element-wise multiplication with their unimodal features:
$$f_{RGB} = \mathbf{I}_{RGB} \odot F_{RGB}, \; f_{Depth} = \mathbf{I}_{Depth} \odot F_{Depth}, $$
herein, $\odot$ is element-wise multiplication. Further, at each level, the feature maps are fused as $f_{fuse} = f_{RGB} \odot f_{Depth}$.

Next, a CA attention mechanism is used to highlight the complementary regions of the two feature maps:
$$\alpha_{RGB} = \sigma(\mathbf{W}_{RGB} \ast f_{RGB}), \; \alpha_{Depth} = \sigma(\mathbf{W}_{Depth} \ast f_{Depth}),$$
$$\alpha_{fuse} = \alpha_{RGB} \odot \alpha_{Depth} \odot \sigma(\mathbf{W}_{fuse} \ast f_{fuse}),$$
$$\alpha = \frac{\alpha_{fuse}}{\alpha_{RGB} + \alpha_{Depth} - \alpha_{fuse}},$$
where $\ast$ denotes the convolution operation, $\mathbf{W}_{RGB}$,  $\mathbf{W}_{Depth}$ and $\mathbf{W}_{fuse}$ are learnable convolutional filters, $\sigma$ is the sigmoid activation function, and $\alpha_{RGB}$, $\alpha_{Depth}$, $\alpha_{fuse}$ are the attention maps for the RGB, Depth and fused feature maps, respectively, the attention map $\alpha$ is the normalized attention map. $\alpha$ provides a material prediction map for the given scene, which we use to create the bounding box with labels $l_i \in \mathcal{L}$ on the RGB map for the given label set $\mathcal{L}$.

\subsection{Visual SLAM ($\mathtt{VSLAM}$) Pipeline }
\label{sec:slam}
We build our 3D semantic mapping on top of a visual SLAM solution. We employ the ORB-SLAM2 \cite{murORB2} algorithm for VSLAM due to its real-time performance, scalability, and portability advantages. Furthermore, ORB-SLAM2 is designed to handle monocular, stereo, and RGB-D cameras, and it is based on the ORB (Oriented FAST and Rotated BRIEF) feature descriptors and uses a combination of point features and line features for robust and accurate localization and mapping, producing sparse 3D reconstruction. In our pipeline, the SLAM module uses the RGB-D data to obtain the point cloud map of the environment in real time. 

\subsection{Voxel Based Matching ($\mathtt{VOXM}$) and 3D Clustering}
\label{sec:mscc}

The voxel-based point cloud matching component is crucial in our architecture. This module uses the depth information from the RGB-D sensor to estimate the 3D coordinates of the bounding boxes obtained through the $\mathtt{OBJ}$ module in the camera coordinate system. Note that both the object detections and the SLAM algorithm's outputs are in the same coordinate system as the RGB-D camera frame. 
Specifically, this component creates a voxel grid to divide a point cloud into smaller voxels during voxelization. It associates each point in the point cloud with the voxel to which it belongs. Then, it iterates over all the voxels, and for each voxel, it finds the closest bounding box. This step is essential as it enables the efficient processing of the point cloud by reducing the number of points that need to be considered for segmentation and material label propagation.

Let $p_i \in \mathcal{P}$ be the set of points in the point cloud (obtained from $\mathtt{VSLAM}$, and $b_i \in \mathcal{B}$ be the set of 3D bounding boxes obtained from $\mathtt{OBJ}$, and $l_i \in \mathcal{L}$ be the label of classified material obtained through $\mathtt{MCN}$. The Voxel-Based Matching Component aims to find a mapping between $b_i$, $l_i$, and the voxels of $p_i$.
Let $\mathcal{V}$ be the voxel grid created by dividing $p_i$ into a 3D grid of small cubes or voxels. Each voxel is assigned a 3D coordinate and contains a set of points from the point cloud that fall within its boundaries.

The mapping between $b_i$ and $\mathcal{V}$ can be represented as a function $\mathcal{M} : b_i \rightarrow \mathcal{V}$, which maps each bounding box in $b_i$ to the voxel in $\mathcal{V}$ that contains the majority of its points.  Once the voxel grid is created, point cloud segmentation is applied in such a way that a static color map $\mathcal{L} \rightarrow \mathcal{C}$ is used to color the material associated with each classified material label in the output semantic map $\mathcal{SM}$ to separate it into different cluster points.
To achieve this objective, we leverage the multi-scale connected components (MSCC) algorithm \cite{2022ISPPCS_MultiScale}, which provides an efficient way to propagate material labels in the point cloud.
The MSCC algorithm is applied on the point cloud after it has been matched with the 3D bounding boxes obtained from object detection using the voxel-based matching component. The algorithm is applied at multiple scales, starting from a large scale and gradually reducing it to capture more fine-grained details in the point cloud. At each scale, the algorithm applies connected component labeling to group points that are spatially close and similar. These groups of points are then assigned a unique label, and the scale is reduced until the minimum scale is reached.

Let $S = {s_1, s_2, \ldots, s_n}$ be the set of scales.
At each scale, $s_i$, the MSCC algorithm performs the following steps:
\begin{enumerate}
    \item Divide $\mathcal{V}$ into a set of larger voxels at scale $s_i$.
    \item Apply connected component labeling to group points that are spatially close and similar to one another within each larger voxel.
    \item Merge the resulting clusters across adjacent voxels, considering their spatial proximity and similarity.
    \item Assign a unique label to each resulting cluster.
\end{enumerate}

The resulting segmentation at scale $s_i$ can be represented as a function $\text{Seg}(s_i): \mathcal{V} \rightarrow \mathcal{L}$, which maps each voxel in $V$ to its assigned label in the segmentation at scale $s_i$. The final segmentation of $\mathcal{V}$ is obtained by merging the segmentations at all scales: $\text{Seg} = \text{merge}(\text{Seg}(s_1), \text{Seg}(s_2), \ldots, \text{Seg}(s_n))$. Where $\text{merge}$ is a function that merges the labels of overlapping voxels across scales. 

Next, the material labels $l_i$ obtained from CAFN are propagated to each cluster obtained from the MSCC algorithm. This is done by finding the closest bounding box for each cluster and assigning the material label of that bounding box to the cluster. This step is important as it enables the creation of a point cloud map with material labels.

This process can be repeated for all segments in the point cloud, resulting in the accurate propagation of material labels to the corresponding segments in the point cloud. The output of the voxel-based matching component is a point cloud map with material labels, which can be used for various applications such as robot navigation, object recognition, and scene understanding. 
The final output is a semantic map $\mathcal{SM}$ represented as a 3D point cloud, where each point is associated with semantic (object type) and material labels. This approach provides a detailed and informative perception of the environment, which is crucial for efficient robot navigation and interaction with its physical surroundings.

\section{Experimental Validation}
\label{experiments}
We train and evaluate our approach using publicly available RGB and RGB-D datasets and demonstrate the accuracy of the semantic map through real-world mobile robot experiments in our lab setting.
We compare various components of our material classification and clustering pipeline with relevant SOTA methods from the literature.

\subsection{Datasets and Model Training} 
Given the fact that most object and material classification datasets available are RGB images, we used the benchmark MINC-2500 \cite{bellUSB2014dataset} (RGB) and Flickr Material Database (FMD) \cite{sharan2009material} (RGB). For RGB and depth network fusion, we used the Washington RGB-D \cite{Lai2011ALH} dataset. 
In our material classification experiments, we grouped the objects together based on their material type. We categorized them into ten material classes: \{\textit{Cardboard, Ceramic, Cloth, Glass, Metal, Paper, Plastic, Rubber, Sponge, Wood}\}. The material type predictions with max probability less than 0.5 are considered as an additional "\textit{other}" type.
Furthermore, for the semantic mapping objective, we used the TUM RGB-D dataset \cite{keimel2012tum}, which is a comprehensive collection of 39 real-world indoor sequences categorized into Handheld SLAM, Robot SLAM, and Dynamic Objects. 
The YOLOv5 network for object detection has been pre-trained on the COCO dataset \cite{lin2014microsoft} and can distinguish between 80 different classes of objects. 
In addition, we created a custom YOLOv5 model built on top of the pre-trained model by adding new object classes, such as board, door, mat, robot, and trash bin, that are not commonly available in existing datasets (but useful in robotics and navigation context \cite{Martins2020ExtendingMW}). We collected 500 images from the public internet for each of these additional classes labeled using the LabelImg annotation tool. 
To verify the effectiveness of our object detection pipeline on these new classes, we compared the "door" and "trash bin" object detections with \cite{Martins2020ExtendingMW} in two of their datasets (sequence1-Kinect and sequence3-astra), where we found our pipeline providing superior detection accuracies (e.g., ours 85.9 \% compared to their's 53.3\% for "door" object detections).

We performed our experiments on a PC with 3 NVIDIA GeForce GPUs. 
The learning rate, weight decay, and mini-batch size are set to 1e-5, 0.0005, and 4, respectively. The training procedure used 50 epochs.
We evaluate our method on the material classification task using five cross-validation splits. Each split consists of roughly 12,440 training images and 3,920 images for testing. During the test, the task of the model is to assign the correct class label to a previously unseen object instance. 
During the inference step in our pipeline applied to real-time RGB-D data, the profiling of average per-keyframe processing times at various stages is: \{$\mathtt{OBJ}$: 56 ms, $\mathtt{MCN}$: 3.27 ms, $\mathtt{VSLAM}$: 6.34 ms, $\mathtt{VOXM}$: 92 ms, Final label propagation: 8.26 ms, Total: 165.87 ms\}. As expected, point cloud segmentation along with YOLOv5 takes significant time, both of which can be upgraded or improved for strict real-time applications.

\subsection{Material Classification}

First, we validate the performance of our $\mathtt{MCN}$ against GoogLeNet \cite{Szegedy2014GoingDW} and VGG\textunderscore CNN\textunderscore M \cite{Kalliatakis2017EvaluatingDC} networks on the benchmark MINC-2500 \cite{bellUSB2014dataset} and FMD \cite{sharan2009material} datasets, based only on the RGB images. The comparison of the accuracy results for the common materials (available in the specific dataset) can be seen in Table~\ref{tab:material_classification_comparison}. 
Our approach demonstrates robust and competitive performance in material classification, excelling particularly in recognizing cloth, plastic, and wood across different RGB datasets.

\begin{figure}
    \centering
    \vspace{-10mm}
    \includegraphics[width=1.15\linewidth]{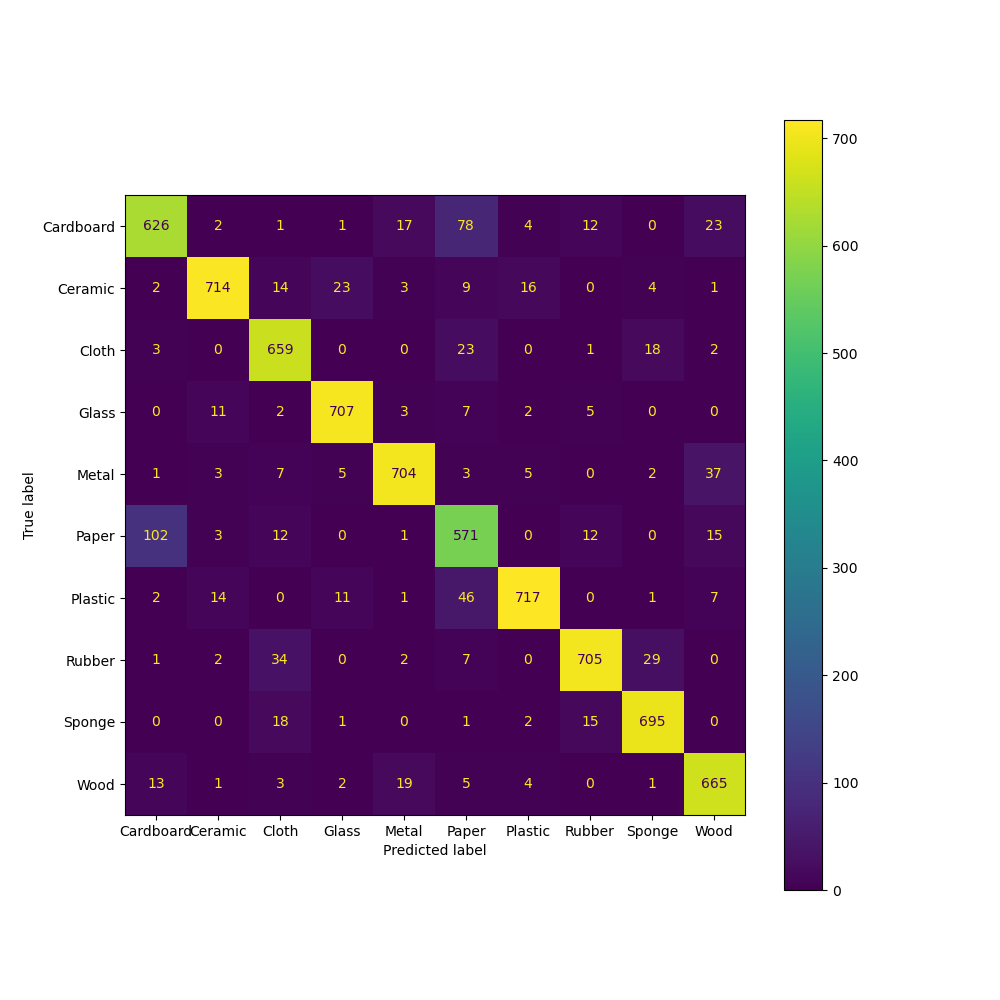}
    \vspace{-12mm}
    \caption{Confusion Matrix of Material Classification Network}
    \label{fig:Conf_Mat}
\end{figure}

\begin{table}[th]
\caption{Comparative analysis of the material classification accuracy (\%) on the RGB datasets.}
\label{tab:material_classification_comparison}
\centering
\resizebox{\linewidth}{!}{
\begin{tabular}{|c|p{2cm}|p{1.2cm}|p{0.5cm}|p{2cm}|p{1.2cm}|p{0.5cm}|}
    \hline
    \multirow{2}{*}{Material} & \multicolumn{3}{c|}{\textbf{MINC-2500 Dataset} \cite{bellUSB2014dataset}} & \multicolumn{3}{c|}{\textbf{FMD Dataset} \cite{sharan2009material}} \\
    \cline{2-7}
    & \textbf{GoogLeNet} \cite{Szegedy2014GoingDW} & \textbf{VGG} \cite{Kalliatakis2017EvaluatingDC} & \textbf{Ours} & \textbf{GoogLeNet} \cite{Szegedy2014GoingDW} & \textbf{VGG} \cite{Kalliatakis2017EvaluatingDC} & \textbf{Ours}\\
    \hline
    Ceramic & \textbf{89.49} & 88.7 & 87.1 & N/A & N/A & N/A \\
    \hline
    Cloth & 81.39 & 79.85 & \textbf{82.6} & 71.07 & 68 & \textbf{84.35} \\
    \hline
    Glass & 84.97 & 84.76 & \textbf{86.3} & 94.03 & \textbf{96.54} & 89.2 \\
    \hline
    Metal & 86.51 & 84.76 & \textbf{87.23} & 86.3 & \textbf{92.28} & 88.4 \\
    \hline
    Paper & \textbf{92.87} & 90.15 & 74.26  & \textbf{90.57} & 82.69 & 77.3 \\
    \hline
    Plastic & 71.04 & 61.39 & \textbf{77.82} & 87.25 & 87.01 & \textbf{89.46} \\
    \hline
    Wood & \textbf{92.52} & 86.47 & 86.38 & 86.87 & 85.35 & \textbf{88.52} \\
    \hline
\end{tabular}
}
\end{table}

Next, we present the accuracy of our material classification network on the Washington RGB-D dataset \cite{Lai2011ALH}, with comparisons to the late fusion scheme~\cite{ding2019DepthAware} as well as schemes that use single-modality (RGB or Depth) inputs. Results in Table~\ref{tab:late_vs_ca} show that our multi-modal CA fusion network outperforms the late fusion scheme with up to 12\% improvement in accuracy over all material classes. The late fusion scheme fails to utilize the depth data effectively and, in some cases, performs more poorly than the RGB-only models. In contrast, the CA fusion effectively fused diverse and contrastive features from both the depth and RGB images and thus provides superior accuracy in all classes. Also, we can observe the advantages of adding the depth modality, which provided consistent improvements in the classification accuracy by our CA fusion method on the RGB-D data compared to the RGB-only modality.

\begin{figure*}[htp]
    \centering
    \begin{subfigure}[b]{0.48\linewidth}
        \centering
        \includegraphics[width=\textwidth]{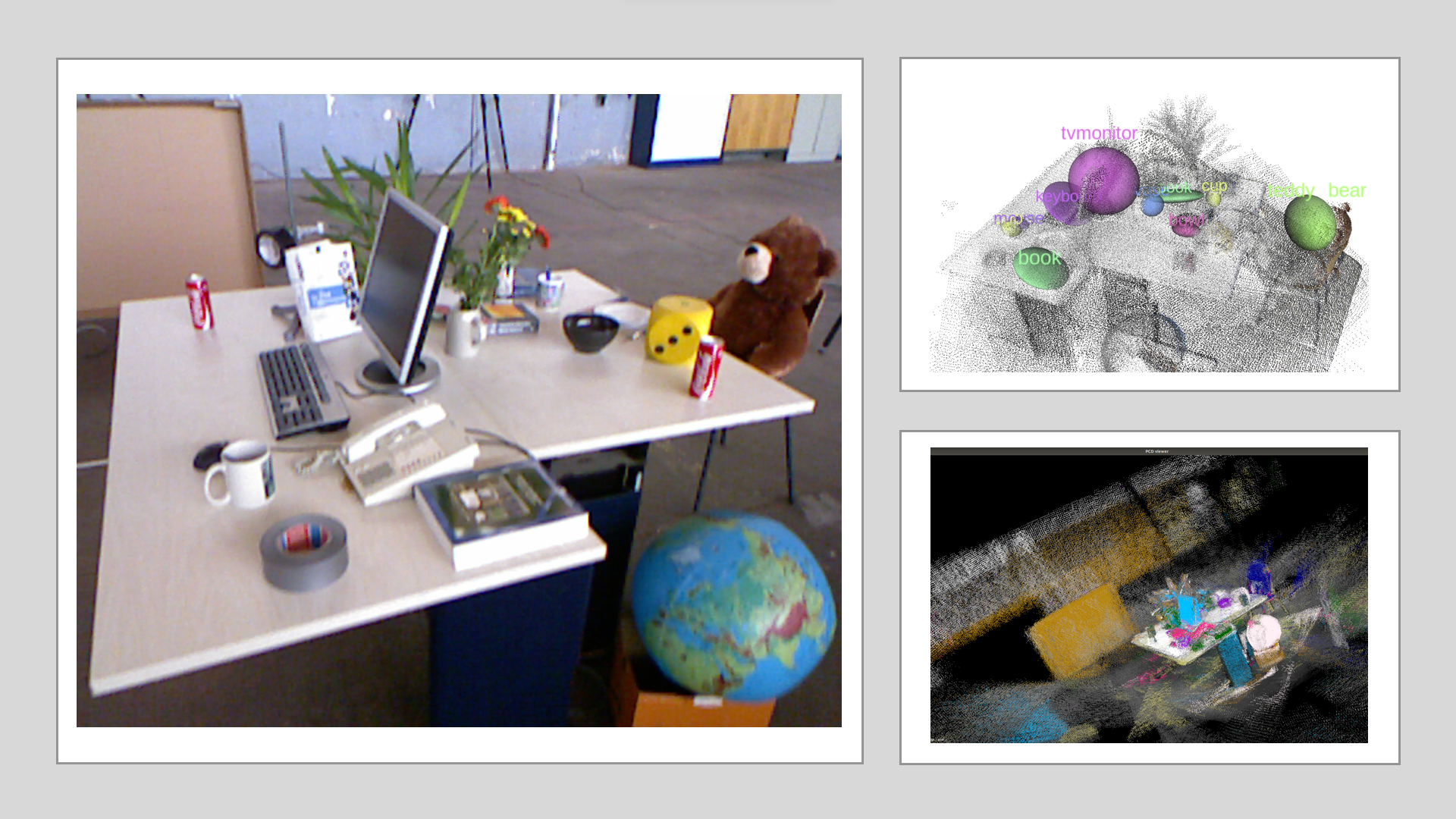}
        \caption{fr2\textunderscore desk sequence.}
        \label{fig:fr2_desk}
    \end{subfigure}
    \begin{subfigure}[b]{0.48\linewidth}
        \centering
        \includegraphics[width=\textwidth]{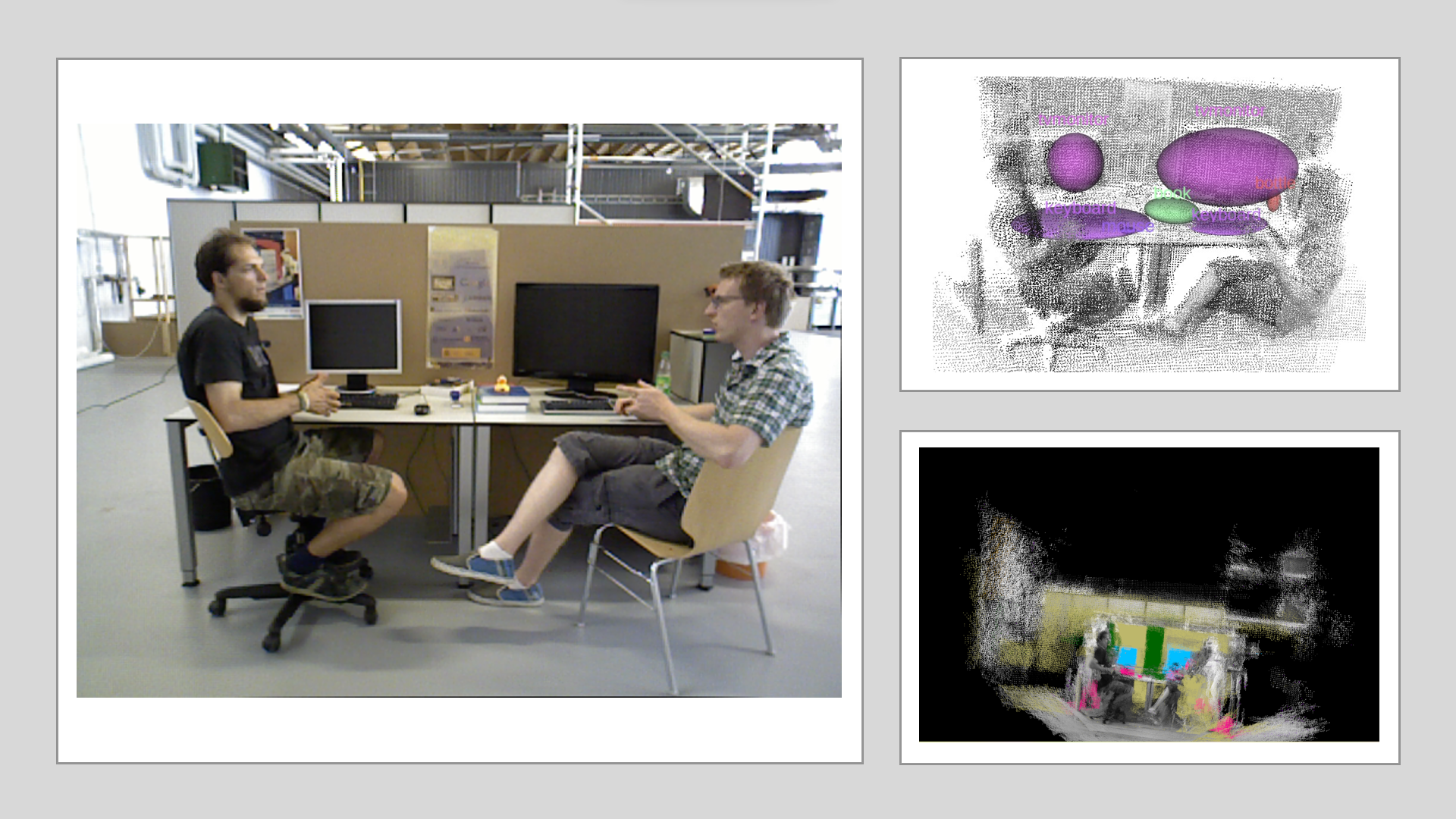}
        \caption{fr3\textunderscore sitting\textunderscore xyz sequence.}
        \label{fig:fr3_sit}
    \end{subfigure}
    \caption{RGB Image, point cloud, and semantic material map comparison with \cite{hempel2022online} (bottom colored point cloud) on two different sequences and their corresponding color representations in the TUM RGB-D dataset \cite{keimel2012tum}: (a) fr2\textunderscore desk sequence, (b) fr3\textunderscore sitting\textunderscore xyz sequence. The colored labels of the respective material classes are shown in Fig.~\ref{fig:semantic_map}.} 
    \label{fig:tum_map}
\end{figure*}



\begin{table}[ht]
\caption{Accuracy of object-oriented material classification methods on the Washington RGB-D dataset \cite{Lai2011ALH}.}
\label{tab:late_vs_ca}
\centering
\resizebox{\linewidth}{!}
{\begin{tabular}{|c|c|c|p{2cm}|p{2cm}|}
\hline
\multirow{2}{*}{\textbf{Material}} & \multirow{2}{*}{\textbf{RGB Only}} & \multirow{2}{*}{\textbf{Depth Only}} & \multicolumn{2}{|c|}{\textbf{RGB + D Fusion}} \\
\cline{4-5}
& & & {\textbf{Late Fusion \cite{ding2019DepthAware}}} & {\textbf{Ours}} \\ 
\hline
Cardboard & 77.40\% & 73.80\% & 73.10\% & \textbf{83.40}\% \\ \hline
Ceramic   & 87.30\% & 84.10\% & 90.80\% & \textbf{95.20}\% \\ \hline
Cloth     & 82.10\% & 79.80\% & 82.50\% & \textbf{87.80}\% \\ \hline
Glass     & 86.30\% & 83.70\% & 86.20\% & \textbf{94.20}\% \\ \hline
Metal     & 87.30\% & 85.30\% & 89.10\% & \textbf{93.80}\% \\ \hline
Paper     & 69.90\% & 68.30\% & 64.20\% & \textbf{76.10}\% \\ \hline
Plastic   & 90.00\% & 87.30\% & 91.20\% & \textbf{95.60}\% \\ \hline
Rubber    & 86.30\% & 85.40\% & 87.20\% & \textbf{94.00}\% \\ \hline
Sponge    & 84.80\% & 82.50\% & 86.50\% & \textbf{92.70}\% \\ \hline
Wood      & 84.30\% & 82.60\% & 83.50\% & \textbf{88.70}\% \\ \hline
\end{tabular}}
\end{table}

To provide more insights, the confusion matrix for our material classification network ($\mathtt{MCN}$) applied on the Washington RGB-D dataset is shown in Fig.~\ref{fig:Conf_Mat}. 
Cascading of multi-level features successively without intermediate level-wise supervision results in ambiguous multi-level combinations, which we see in the results for the late fusion network. Here, the high-level contexts are not well incorporated into shallow layers. Adding intermediate supervision, the multi-modal fusion network can learn level-specific predictions, as we observe in our CA fusion network. Visually, the shallow layers can identify edge information, and the deep layers can learn global contexts to detect objects accurately. The CAFN involves cross-modal residual connections and complementarity-aware supervisions, captures the cooperated information, and boosts better cross-modal combinations, thus generating more accurate object detection and classification. The material classification results confirm that the model benefits from the more sufficient fusion of multi-modal and multi-level features, and hence CA fusion models achieve much better performance than the late fusion models.

\subsection{Point Cloud Segmentation and 3D Clustering}
\label{sec:clustering}
We used the TUM RGB-D dataset \cite{keimel2012tum} to evaluate the performance of our 3D clustering of material labels with the SLAM integration. Multiple objects are successfully detected and represented as semantic labels on the map with their corresponding material type. 
We were able to estimate the size of objects accurately using their point clouds, which gives an idea of their general dimensions. 
As a preliminary evaluation, we qualitatively compare our work to the closest relevant work \cite{hempel2022online} that performs 3D object segmentation on the point clouds from the SLAM algorithm to refine the localization accuracy. 
The corresponding examples are shown in Fig.~\ref{fig:tum_map}. As we see, the segmentation in \cite{hempel2022online} detects only a few objects and naively clusters them as 3D spheres without consistency checks. In contrast, our pipeline detects various objects and estimates their material type to achieve a smoother and finer clustering consistency across the voxels.

To formally evaluate, we compare our object-oriented clustering results against \cite{OnlineObjSemMap}, where a similar semantic object-level 3D clustering is proposed (without material classification). To allow this comparison, we disregarded material classification and applied the 3D clustering based on the object semantics. Table~\ref{tab:comparison_with_onnline} shows the performance metrics in terms of mean average precision (mAP) (for object/material detection accuracies), Intersection over Union (IoU) (for clustering accuracy), and the number of object detections (for object-oriented effectiveness) on two different TUM dataset sequences. Our model has outperformed \cite{OnlineObjSemMap} in all the metrics due to the robust multiscale clustering. 

\begin{table}[h]
\caption{Results of 3D clustering compared with \cite{OnlineObjSemMap}.}
\label{tab:comparison_with_onnline}
\centering
\resizebox{\linewidth}{!}{
\begin{tabular}{|p{1.8cm}|p{1cm}|p{1.3cm}|p{0.7cm}|p{0.7cm}|p{1.3cm}|p{0.7cm}|p{0.7cm}|}
    \hline
   \textbf{TUM RGB-D} & \multirow{2}{*}{\textbf{\#Objects}} & \multicolumn{3}{c|}{\textbf{3D Object Segmentation \cite{OnlineObjSemMap}}} & \multicolumn{3}{c|}{\textbf{Ours}} \\
    \cline{3-8}
    dataset \cite{keimel2012tum} & & \textbf{\#Detections} & \textbf{IoU} & \textbf{mAP} & \textbf{\#Detections} & \textbf{IoU} & \textbf{mAP} \\
    \hline
    fr2\textunderscore desk & 10 & 581 & 0.567 & 0.671 & \textbf{598} & \textbf{0.776} & \textbf{0.734} \\
    \hline
    fr3\textunderscore sitting\textunderscore xyz & 5 & 356 & 0.615 & 0.652 & \textbf{367} & \textbf{0.765} & \textbf{0.718}\\
    \hline
\end{tabular}
}
\end{table}

\begin{figure*}[t]
    \centering
    \includegraphics[width=0.95\linewidth]{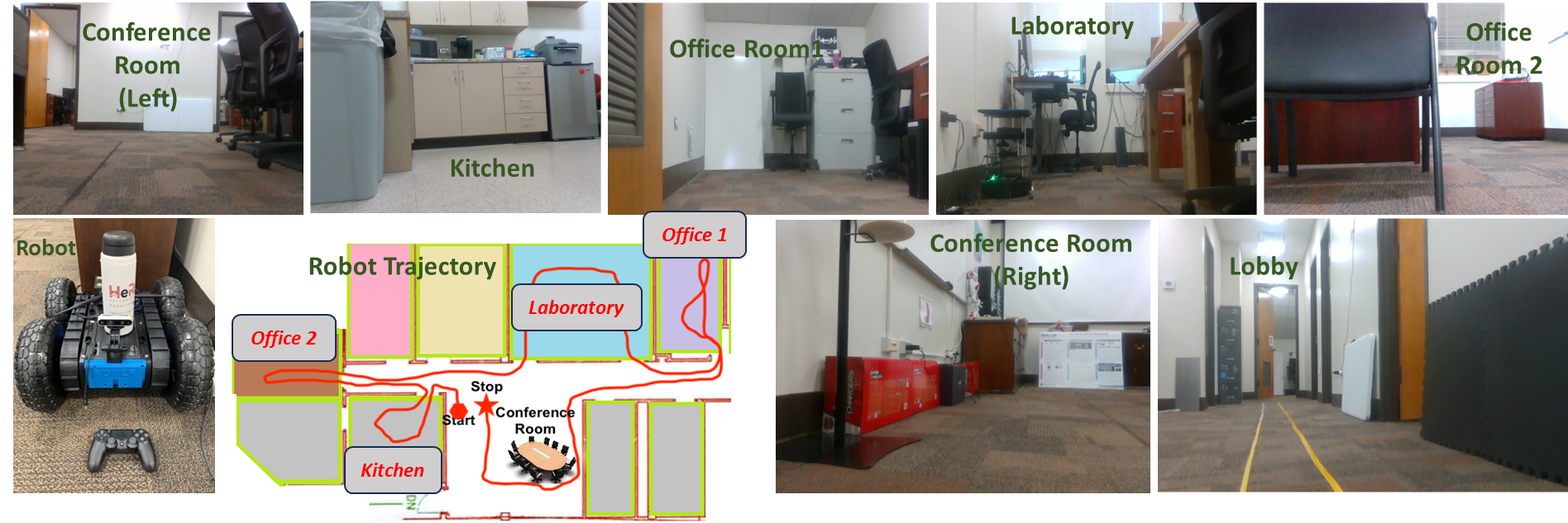}
    \vspace{-2mm}
    \caption{Robot setup for the real-world mobile robot RGB-D experiments and dataset collection.}
    \label{fig:robot-setup}
\end{figure*}

\begin{table*}[ht]
\caption{3D Material Mapping accuracy results for the real-world robot dataset experiments in multiple rooms.}
\label{tab:robot_results}
\centering
\resizebox{\linewidth}{!}{
\begin{tabular}{|c|c|c|c|c|c|c|c|c|c|c|c|c|c|c|c|}
    \hline
     \multirow{2}{*}{\textbf{Material Class}} & 
 \multicolumn{3}{|c|}{\textbf{Kitchen}} & \multicolumn{3}{|c|}{\textbf{Office Room 1}} & \multicolumn{3}{|c|}{\textbf{Office Room 2}} & \multicolumn{3}{|c|}{\textbf{Laboratory Room}} & \multicolumn{3}{|c|}{\textbf{Conference Room}} \\
    \cline{2-16}
   & \textbf{\#Objects} & \textbf{IoU} & \textbf{mAP} & \textbf{\#Objects} & \textbf{IoU} & \textbf{mAP} & \textbf{\#Objects} & \textbf{IoU} & \textbf{mAP} & \textbf{\#Objects} & \textbf{IoU} & \textbf{mAP} & \textbf{\#Objects} & \textbf{IoU} & \textbf{mAP} \\
    \hline
   Cardboard & 4 & 0.674 & 0.652 & 5 & 0.782 & 0.626 & 1 & 0.863 & 0.678 & 4 & 0.768 & 0.61 & 8 & 0.784 & 0.622 \\
    \hline
    Cloth & - & - & - & - & - & - & - & - & - & 1 & 0.712 & 0.666 & 1 & 0.67 & 0.643 \\
    \hline
    Glass & - & - & - & 1 & 0.672 & 0.654 & 1 & 0.858 & 0.676 & 2 & 0.715 & 0.623 & - & - & - \\
    \hline
    Metal & 4 & 0.789 & 0.613 & 9 & 0.816 & 0.586 & 3 & 0.867 & 0.616 & 3 & 0.85 & 0.631 & 2 & 0.732 & 0.66 \\
    \hline
    Paper & 3 & 0.852 & 0.632 & 2 & 0.84 & 0.72 & 2 & 0.788 & 0.66  & 1 & 0.813 & 0.668 & 3 & 0.88 & 0.593 \\
    \hline
    Plastic & 5 & 0.849 & 0.645 & - & - & - & - & - & - & 2 & 0.833 & 0.659 & 1 & 0.872 & 0.646 \\
    \hline
    Rubber & - & - & - & - & - & - & - & - & - & 1 & 0.768 & 0.638 & 1 & 0.776 & 0.631 \\
    \hline
    Wood & 6 & 0.839 & 0.666 & 7 & 0.778 & 0.665 &5 & 0.705 & 0.657 &  8 & 0.845 & 0.659 & 5 & 0.831 & 0.661 \\
    \hline
    Other (e.g., Fiber) & - & - & - & - & - & - & 3 & 0.786 & 0.622 & 4 & 0.816 & 0.586 & 4 & 0.844 & 0.943 \\
   \hline
   Average & 22 (total) & 0.801 & 0.642 & 24 (total) & 0.778 & 0.650 & 15 (total) & 0.811 & 0.652 & 26 (total) & 0.791 & 0.638 & 25 (total) & 0.799 & 0.675 \\
    \hline
\end{tabular}
}
\end{table*}

\begin{figure*}[t]
    \centering
    \scriptsize
    \begin{tabular}{cc}
        \hspace{-3mm} 
        {\begin{turn}{90} \hspace{4mm} \bf Conference Room \end{turn}}
&
        \hspace{-4mm}   \includegraphics[width=0.95\linewidth]{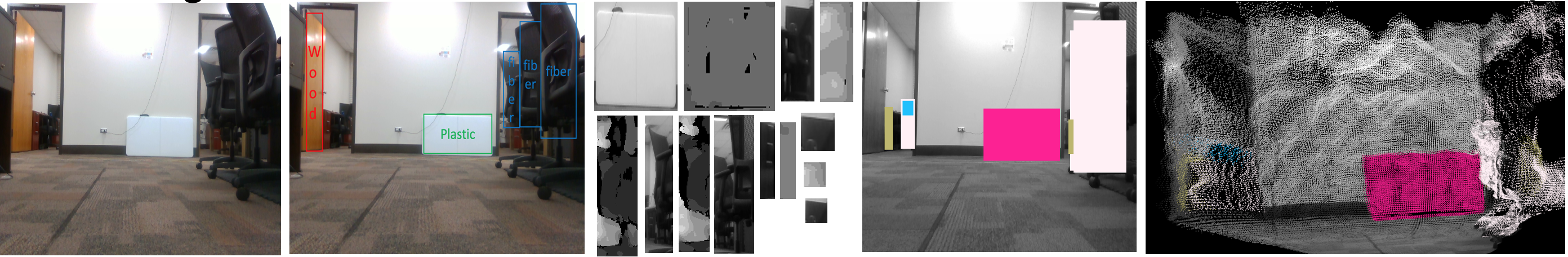}
    \\
    
   \hspace{-3mm}          {\begin{turn}{90} \hspace{8mm} \bf Laboratory \end{turn}}
&
       \hspace{-4mm}        \includegraphics[width=0.95\linewidth]{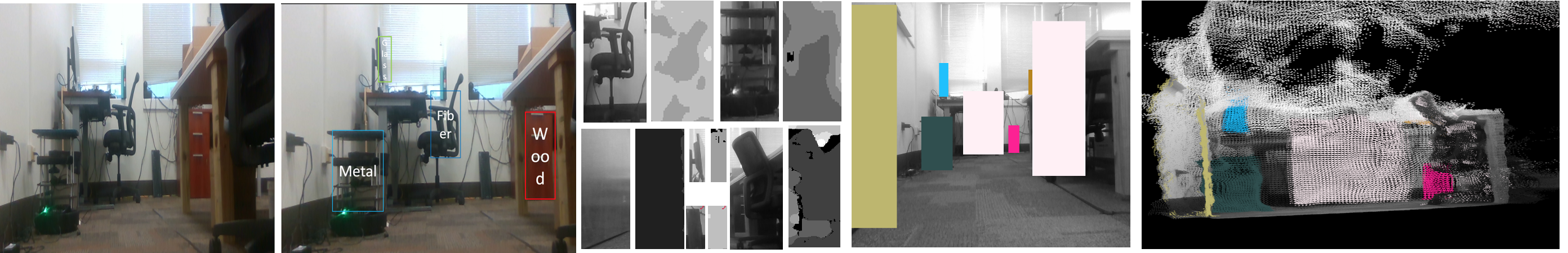}
        \\

    \hspace{-3mm}        {\begin{turn}{90} \hspace{8mm} \bf Office 1 \end{turn}}
&
    \hspace{-4mm}       \includegraphics[width=0.95\linewidth]{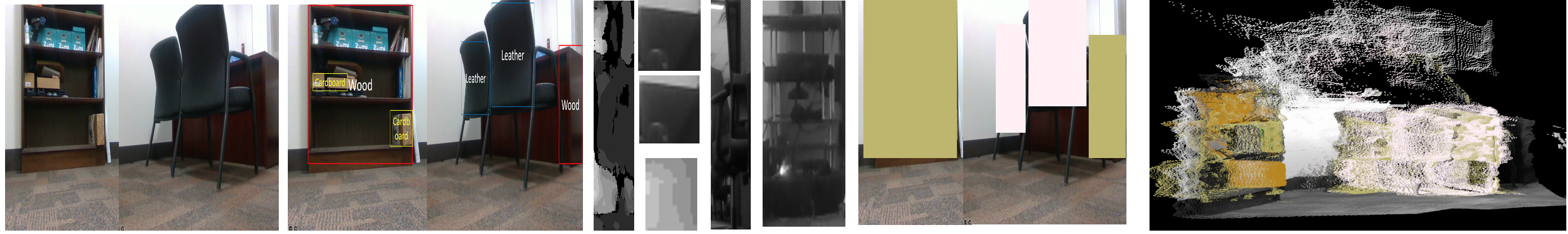}
    \\
    
    \hspace{-3mm}        {\begin{turn}{90} \hspace{8mm} \bf Office 2 \end{turn}}
&
    \hspace{-4mm}       \includegraphics[width=0.95\linewidth]{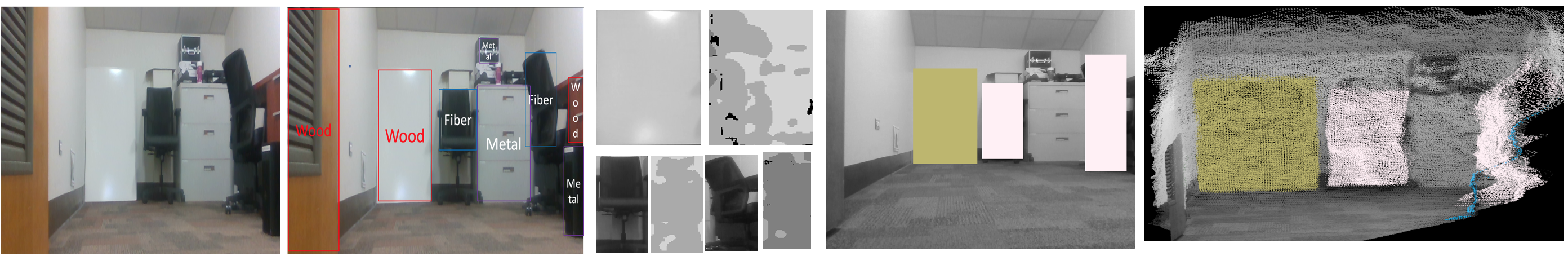}
    \\
    \hspace{-3mm}       {\begin{turn}{90} \hspace{8mm} \bf Kitchen \end{turn}}
&
        \hspace{-4mm}   \includegraphics[width=0.95\linewidth]{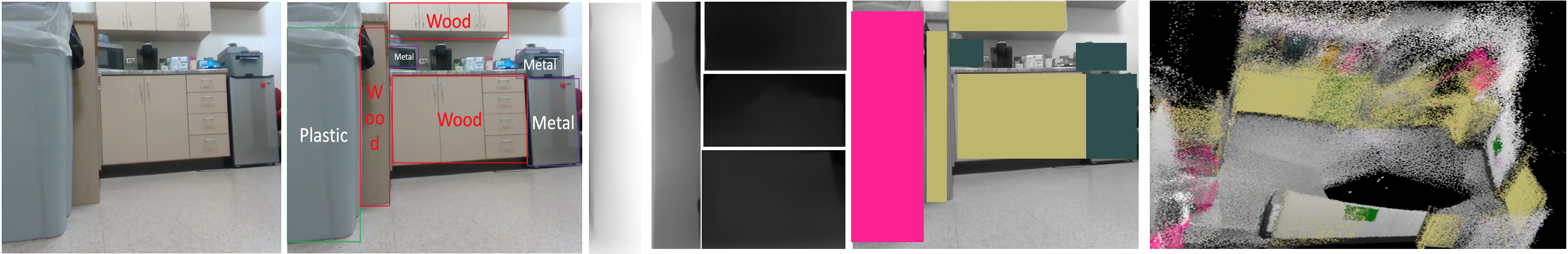}    
    \\
   \end{tabular}
    \caption{Demonstration of sample outputs at various stages in our pipeline for the real-world robot RGB-D datasets. From left: RGB Image, ground truth material labels, segmented objects, material classification, and the final 3D clustering.
}
    \label{fig:conslidated_image}
    \vspace{-2mm}
\end{figure*} 

\subsection{Real-world Mobile Robot Experiments}
\label{sec:robot-experiments}
We used a mobile robot platform (4WD Rover Zero V3) equipped with a D435i Intel Realsense RGB-D camera, an RPLidar-v3, and an NVIDIA Jetson Nano. 
We conducted teleoperated mobile robot trajectories in multiple rooms (consisting of diverse scenes and different object compositions) in controlled indoor conditions and recorded them as ROSbag datasets.
The experiments were run on an Intel Core i7 laptop with Nvidia GeForce GTX 1050 Ti running on Ubuntu 18.04 with ROS-Melodic. The goal was to obtain enhanced 3D maps with object material information that does not change over time, such as doors, desks, chairs, and other objects. 
We intentionally prepared the settings such that the datasets consist of multiple objects made of similar materials (e.g., cabinets, doors, desks made of wood) and the same object types made of different materials (e.g., wooden, plastic, and metal chairs).
A manually labeled ground truth map specified the static objects' location.
Each sequence included raw data from multiple sources, including two RGB-D cameras, LiDAR, and odometry. Fig.~\ref{fig:robot-setup} shows the robot setup, experiment trajectory, and visuals for each room.
Appendix A provides more details on the objects and material information of different rooms in the dataset.

We organized the results based on the material types. Fig.~\ref{fig:conslidated_image} showcases example outputs from multiple stages in our object-oriented pipeline, and Table~\ref{tab:robot_results} shows the comprehensive accuracy results in different rooms along with the number of objects representing each material type in each room as per the ground truth.
As we can observe, the proposed approach has effectively clustered the material properties in the map with acceptable classification and 3D localization accuracies (approx. 0.8 mean IoU and 0.65 mAP on average).
This is an acceptable and significant result (e.g., the mean IoU reported in \cite{Zhao2020SimultaneousMS} for 3D reconstruction is 0.39 for an industrial scenario).
The attached video provides a demonstration and in-depth information on these results.

\section{Conclusion}
We proposed an object-oriented pipeline for 3D semantic mapping of surface material properties in a mobile robotics environment. The proposed approach significantly advances the SOTA in mapping and perception by integrating semantic objects and material identification into a cohesive and effective mapping system. The resulting semantic map associates each point in a 3D point cloud with semantic and material labels.
Our extensive experiments, conducted with public and in-house datasets, have demonstrated the robustness and accuracy of our method in creating detailed semantic maps better than comparable approaches. These maps have been validated with qualitative results, showing successful object detection and material classification, which are crucial for robots operating in dynamic and unknown environments.
This dual-label (object and material) mapping can prove to be a valuable asset for static landmark identification, facilitating more precise trajectory planning.

\bibliography{ref}
\bibliographystyle{IEEEtran}

\clearpage

\appendix{Additional Information on the Real-World Mobile Robot Material RGB-D Dataset with Extended Results}

\begin{table}[htp]
\centering
\begin{tabular}{|l|l|l|}
    \hline
    \multicolumn{3}{|c|}{\textbf{Conference Room}} \\
    \hline
    \textbf{Object} & \textbf{Material Type} & \textbf{Count} \\
    \hline
    cloth sheet	& cloth	& 1 \\
    \hline
    cardboard boxes	& cardboard	& 8 \\
    \hline
    chair & fiber & 3 \\
    \hline
    door & wood & 2 \\
    \hline
    desks & wood & 3 \\
    \hline
    mat & rubber & 1 \\
    \hline
    plastic board & plastic & 1 \\
    \hline
    screen & polyester & 1 \\
    \hline
    posters & paper & 3 \\
    \hline
    robots & metal & 2 \\
    \hline
\end{tabular}
\caption{List of Objects and their Material Type in Conference Room.}
\end{table}

\begin{table}[htp]
\centering
\begin{tabular}{|l|l|l|}
    \hline
    \multicolumn{3}{|c|}{\textbf{Laboratory Room}} \\
    \hline
    \textbf{Object} & \textbf{Material Type} & \textbf{Count} \\
    \hline
    cardboard boxes	& cardboard	& 4 \\
    \hline
    cloth sheet	& cloth	& 1 \\
    \hline
    chair & fiber & 4 \\
    \hline
    door & wood & 3 \\
    \hline
    desks & wood & 4 \\
    \hline
    keyboard & plastic & 1 \\
    \hline
    mat & rubber & 1 \\
    \hline
    metal desk & metal & 1 \\
    \hline
    metal plate & metal & 1 \\
    \hline
    monitor & glass & 2 \\
    \hline
    plastic board & plastic	& 1 \\
    \hline
    white board & wood & 1 \\
    \hline
    posters & paper & 1 \\
    \hline
    robot & metal & 1 \\
    \hline
\end{tabular}
\caption{List of Objects and their Material Type in the Laboratory Room.}
\end{table}

\begin{table}[htp]
\centering
\begin{tabular}{|l|l|l|}
    \hline
    \multicolumn{3}{|c|}{\textbf{Kitchen Room}} \\
    \hline
    \textbf{Object} & \textbf{Material Type} & \textbf{Count} \\
    \hline
    bottles	& plastic & 2 \\
    \hline
    cardboard boxes	& cardboard	& 4 \\
    \hline
    coffee machine & metal	& 1 \\
    \hline
    cupboard & wood	& 3 \\
    \hline
    door & wood	& 2 \\
    \hline
    desks & wood & 1 \\
    \hline
    microwave & metal & 1 \\
    \hline
    printer	& metal	& 1 \\
    \hline
    plastic board & plastic	& 1 \\
    \hline
    refrigerator & metal & 1 \\
    \hline
    trash bin & plastic	& 2 \\
    \hline
    posters	& paper	& 3 \\
    \hline
\end{tabular}
\caption{List of Objects and their Material Type in Kitchen Room.}
\end{table}

\begin{table}[htp]
\centering
\begin{tabular}{|l|l|l|}
    \hline
    \multicolumn{3}{|c|}{\textbf{Office Room1}} \\
    \hline
    \textbf{Object} & \textbf{Material Type} & \textbf{Count} \\
    \hline
    cardboard boxes	& cardboard	& 5 \\
    \hline
    chair & metal & 3 \\
    \hline
    door & wood & 2 \\
    \hline
    desks & wood & 5 \\
    \hline
    monitor	& glass	& 1 \\
    \hline
    robot & metal & 6 \\
    \hline
    posters & paper & 2 \\
    \hline
\end{tabular}
\caption{List of Objects and their Material Type in Office Room 1.}
\end{table}

\begin{table}[htp]
\centering
\begin{tabular}{|l|l|l|}
    \hline
    \multicolumn{3}{|c|}{\textbf{Office Room 2}} \\
    \hline
    \textbf{Object} & \textbf{Material Type} & \textbf{Count} \\
    \hline
    cardboard box & cardboard & 1 \\
    \hline
    chair & fiber & 3 \\
    \hline
    cupboard & metal & 1 \\
    \hline
    door & wood & 2 \\
    \hline
    desks & wood & 2 \\
    \hline
    monitor & glass & 1 \\
    \hline
    metal desk & metal & 1 \\
    \hline
    cpu	& metal	& 1 \\
    \hline
    white board	& wood & 1 \\
    \hline
    poster & paper & 2 \\
    \hline
\end{tabular}
\caption{List of Objects and their Material Type in Office Room 2.}
\end{table}

\begin{table*}[h]
\centering
\begin{tabular}{|l|l|l|l|l|l|l|}
    \hline
    \multicolumn{7}{|c|}{\textbf{Conference Room}} \\
    \hline
    \textbf{Object} & \textbf{Detections} & \textbf{Material Prediction} & \textbf{IoU} & \textbf{mAP} & \textbf{TP} & \textbf{FP} \\
    \hline
    cloth sheet & 52 & cloth & 0.67 & 0.643 & 33 & 19 \\
    \hline
    cardboard boxes & 203 & cardboard & 0.784 & 0.622 & 126	& 77 \\
    \hline
    chair & 118 & other & 0.823 & 0.613 & 72 & 46 \\
    \hline
    door & 72 & wood & 0.841 & 0.648 & 47 & 25 \\
    \hline
    desks & 337 & wood & 0.82 & 0.673 & 226 & 111 \\
    \hline
    mat & 155 & rubber & 0.776 & 0.631 & 98 & 57 \\
    \hline
    plastic board & 123 & plastic & 0.872 & 0.646 & 79 & 44 \\
    \hline
    screen & 107 & other & 0.864 & 0.672 & 72 & 35 \\
    \hline
    posters & 145 & paper & 0.88 & 0.593 & 86 & 59 \\
    \hline
    robots & 65 & metal & 0.732 & 0.66 & 43 & 22 \\
    \hline
\end{tabular}
\caption{Conference Room - Object Detections and Material Classification}
\end{table*}

\begin{table*}[h]
\centering
\begin{tabular}{|l|l|l|l|l|l|l|}
    \hline
    \multicolumn{7}{|c|}{\textbf{Kitchen Room}} \\
    \hline
    \textbf{Object} & \textbf{Detections} & \textbf{Material Prediction} & \textbf{IoU} & \textbf{mAP} & \textbf{TP} & \textbf{FP} \\
    \hline
    cardboard boxes	& 120 & cardboard & 0.674 & 0.652 & 78 & 42 \\
    \hline
    cupboard & 220 & wood & 0.838 & 0.646 & 142 & 78 \\
    \hline
    door & 52 & wood & 0.824 & 0.692 & 36 & 16 \\
    \hline
    desks & 43 & wood & 0.856 & 0.66 & 29 & 14 \\
    \hline
    mat	& 4	& rubber & 0.74 & 0.75 & 3 & 1 \\
    \hline
    microwave & 42 & metal & 0.82 & 0.644 & 27 & 15 \\
    \hline
    printer & 8	& metal	& 0.746	& 0.622	& 5	& 3 \\
    \hline
    plastic board & 30 & plastic & 0.822 & 0.632 & 19 & 11 \\
    \hline
    refrigerator & 26 & metal & 0.8 & 0.572 & 15 & 11 \\
    \hline
    trash bin & 24 & plastic & 0.876 & 0.658 & 16 & 8 \\
    \hline
    posters & 62 & paper & 0.852 & 0.632 & 39 & 23 \\
    \hline
\end{tabular}
\caption{Kitchen Room - Object Detections and Material Classification}
\end{table*}

\begin{table*}[h]
\centering
\begin{tabular}{|l|l|l|l|l|l|l|}
    \hline
    \multicolumn{7}{|c|}{\textbf{Office Room 1}} \\
    \hline
    \textbf{Object} & \textbf{Detections} & \textbf{Material Prediction} & \textbf{IoU} & \textbf{mAP} & \textbf{TP} & \textbf{FP} \\
    \hline
    cardboard boxes	& 118 & cardboard & 0.782 & 0.626 & 74 & 44 \\
    \hline
    chair & 26 & metal & 0.867 & 0.532 & 14 & 12 \\
    \hline
    door & 128 & wood & 0.723 & 0.672 & 86 & 42 \\
    \hline
    desks & 134 & wood & 0.833 & 0.658 & 88 & 46 \\
    \hline
    monitor & 43 & glass & 0.672 & 0.654 & 28 & 15 \\
    \hline
    posters & 28 & paper & 0.84 & 0.72 & 20 & 8 \\
    \hline
    robot & 14 & metal & 0.765 & 0.64 & 9 & 5 \\
    \hline
\end{tabular}
\caption{Office Room 1 - Object Detections and Material Classification}
\end{table*}

\begin{table*}[h]
\centering
\begin{tabular}{|l|l|l|l|l|l|l|}
    \hline
    \multicolumn{7}{|c|}{\textbf{Laboratory Room}} \\
    \hline
    \textbf{Object} & \textbf{Detections} & \textbf{Material Prediction} & \textbf{IoU} & \textbf{mAP} & \textbf{TP} & \textbf{FP} \\
    \hline
    cardboard boxes	& 152 & cardboard & 0.768 & 0.61 & 93 & 59 \\
    \hline
    cloth sheet	& 24 & cloth & 0.712 & 0.666 & 16 & 8 \\
    \hline
    chair & 206 & other & 0.816	& 0.586	& 121 & 85 \\
    \hline
    door & 154 & wood & 0.853 & 0.626 & 97 & 57 \\
    \hline
    desks & 208 & wood & 0.81 & 0.674 & 140 & 68 \\ 
    \hline
    keyboard & 98 & plastic & 0.832 & 0.666 & 65 & 33 \\
    \hline
    mat	& 156 & rubber & 0.768 & 0.638 & 99 & 57 \\
    \hline
    metal desk & 72	& metal & 0.845 & 0.62 & 45 & 27 \\
    \hline
    metal plate & 88 & metal & 0.887 & 0.615 & 54 & 34 \\
    \hline
    monitor & 29 & glass & 0.715 & 0.623 & 18 & 11 \\
    \hline
    plastic board & 70 & plastic & 0.834 & 0.652 & 46 & 24 \\
    \hline
    white board & 68 & other & 0.857 & 0.533 & 36 & 32 \\
    \hline
    posters & 20 & paper & 0.813 & 0.668 & 13 & 7 \\
    \hline
    robot & 57 & metal & 0.817 & 0.657 & 38 & 19 \\
    \hline
\end{tabular}
\caption{Laboratory Room - Object Detections and Material Classification}
\end{table*}

\begin{table*}[t]
\centering
\begin{tabular}{|l|l|l|l|l|l|l|}
    \hline
    \multicolumn{7}{|c|}{\textbf{Office Room 2}} \\
    \hline
    \textbf{Object} & \textbf{Detections} & \textbf{Material Prediction} & \textbf{IoU} & \textbf{mAP} & \textbf{TP} & \textbf{FP} \\
    \hline
    cardboard boxes & 72 & cardboard & 0.863 & 0.678 & 49 & 23 \\
    \hline
    chair & 86 & other & 0.786 & 0.622 & 54 & 32 \\
    \hline
    cupboard & 192 & metal & 0.887 & 0.61 & 117 & 75 \\
    \hline
    door & 168 & wood & 0.69 & 0.684 & 115 & 53 \\
    \hline
    desks & 84 & wood & 0.712 & 0.642 & 54 & 30 \\
    \hline
    monitor & 20 & glass & 0.858 & 0.676 & 14 & 6 \\
    \hline
    metal desk & 136 & metal & 0.831 & 0.678 & 92 & 44 \\
    \hline
    cpu	& 54 & metal & 0.882 & 0.56 & 30 & 24 \\
    \hline
    white board & 163 & wood & 0.712 & 0.644 & 105 & 58 \\
    \hline
    poster & 138 & paper & 0.788 & 0.66 & 92 & 46 \\
    \hline
\end{tabular}
\caption{Office Room 2 - Object Detections and Material Classification}
\end{table*}


\begin{figure*}[htp]
    \centering
    \includegraphics[width=16cm]{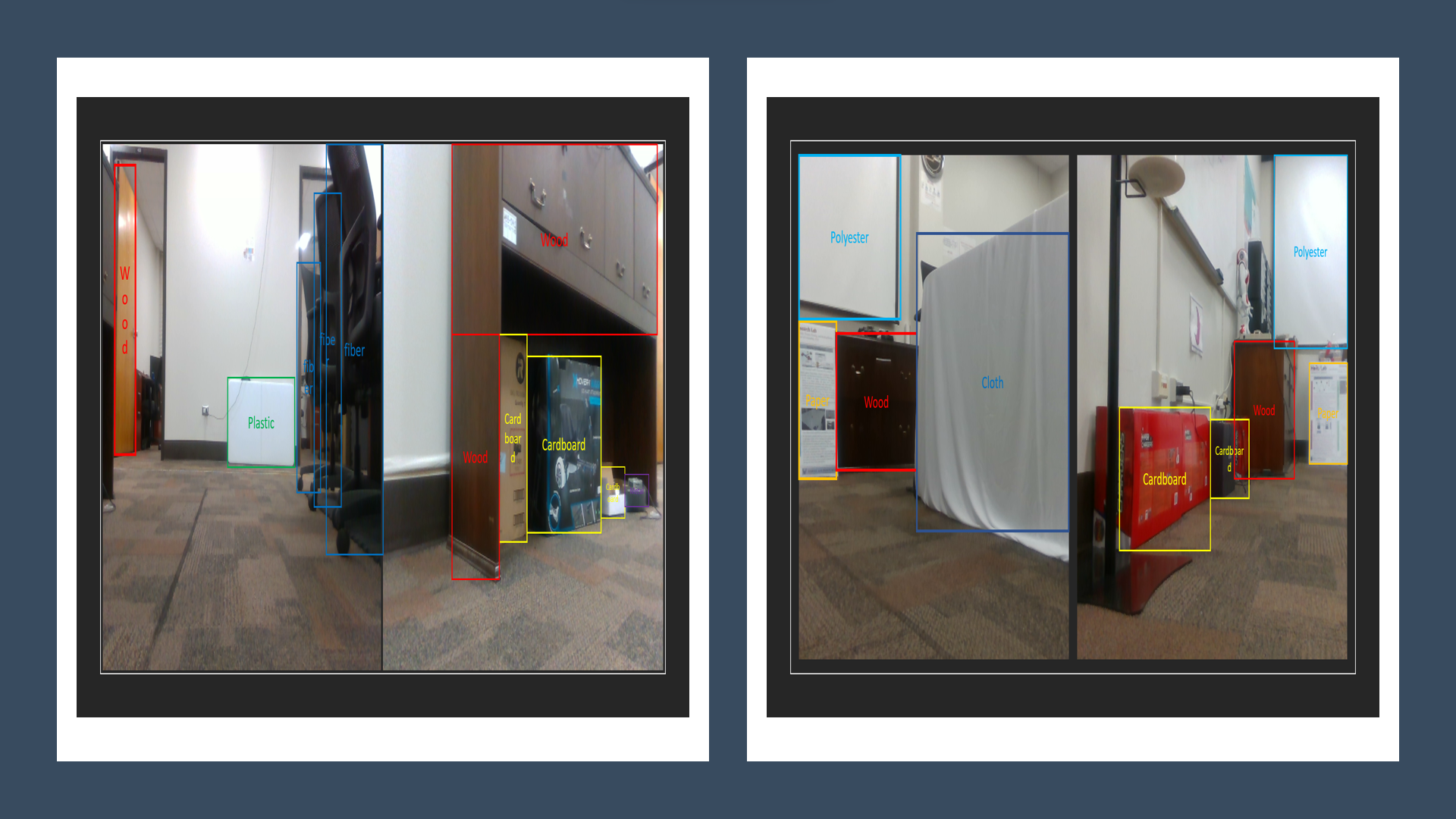}
    \caption{ Conference Room - RGB Images and Ground Truth of Materials}
    \label{fig:conf rgb}
\end{figure*}

\begin{figure*}[htp]
    \centering
    \includegraphics[width=16cm]{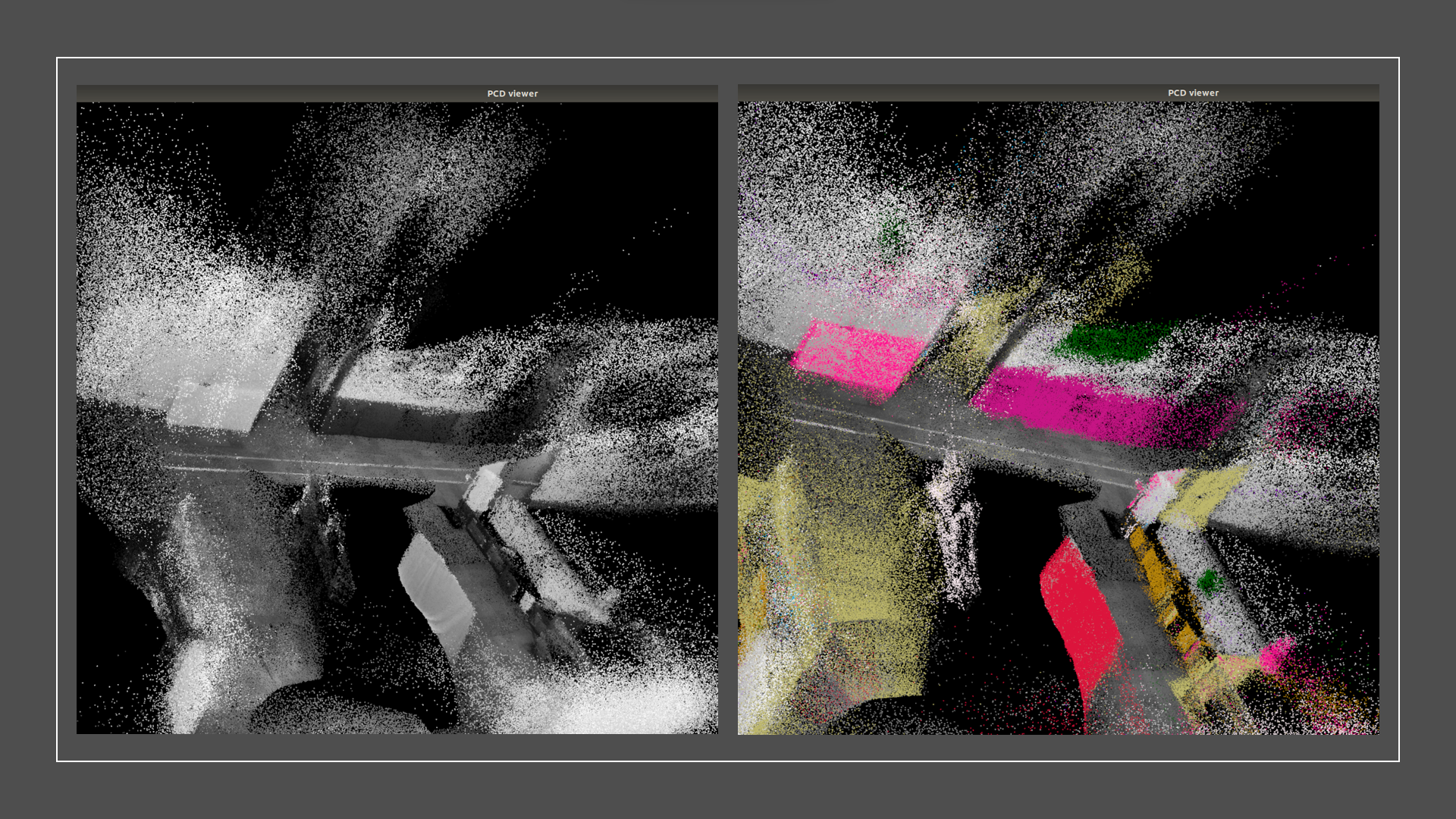}
    \caption{Conference Room - Point Cloud and Semantic Material Map}
    \label{fig:conf room}
\end{figure*}

\begin{figure*}[htp]
    \centering
    \includegraphics[width=16cm]{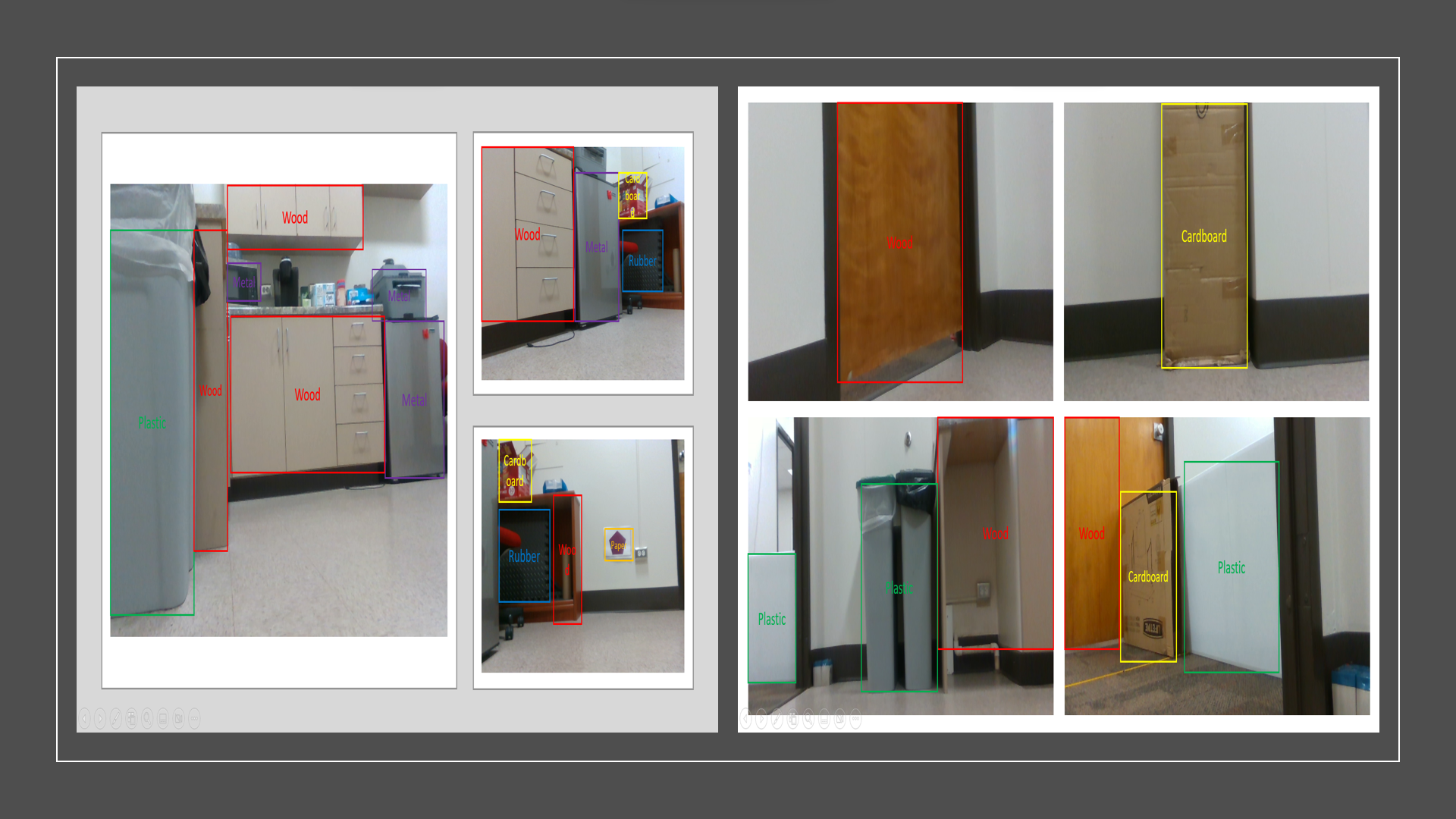}
    \caption{Kitchen - RGB Images and Ground Truth}
    \label{fig:kit rgb}
\end{figure*}

\begin{figure*}[htp]
    \centering
    \includegraphics[width=16cm]{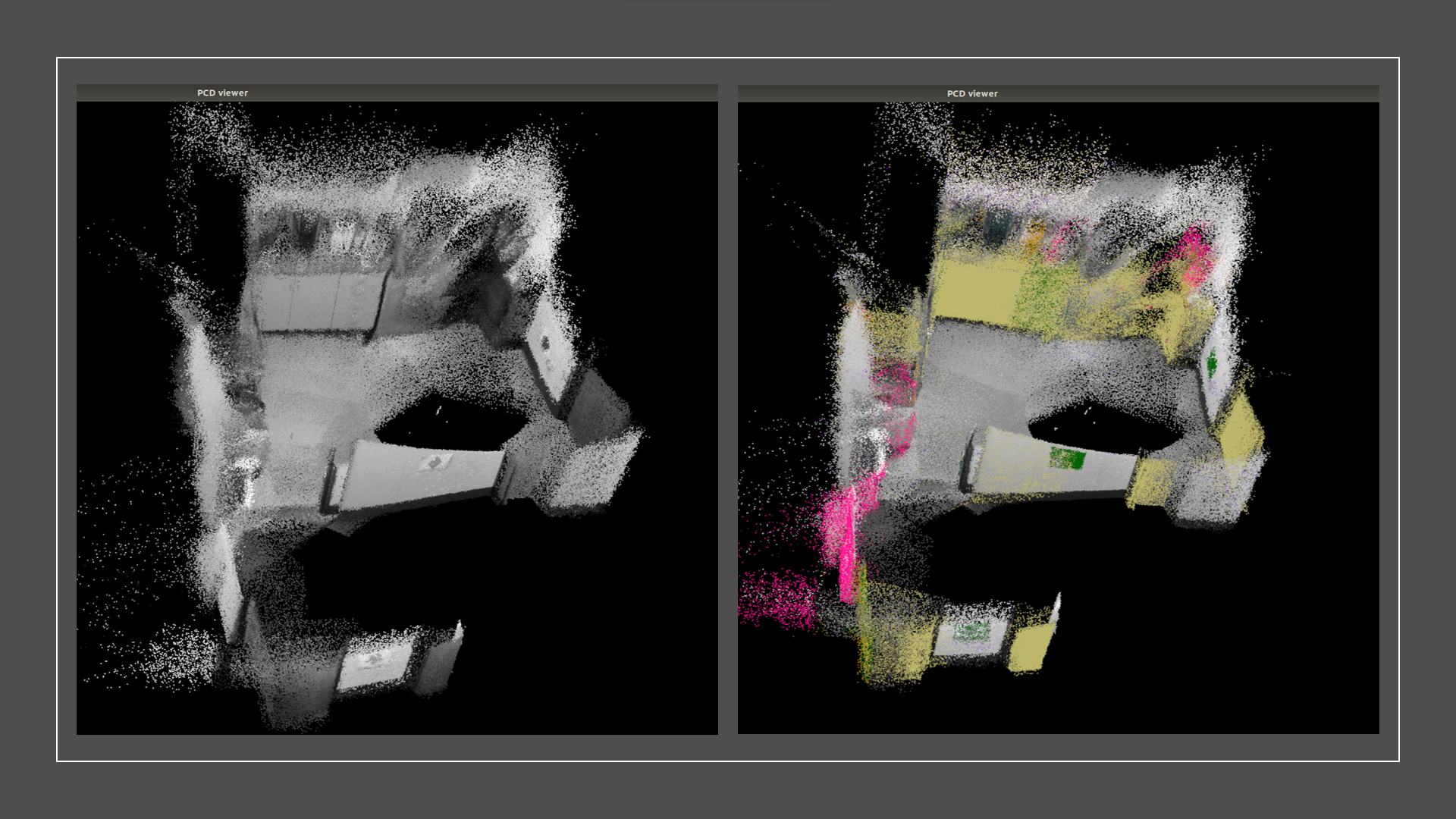}
    \caption{Kitchen - Point Cloud and Semantic Material Map}
    \label{fig:kit room}
\end{figure*}

\begin{figure*}[htp]
    \centering
    \includegraphics[width=16cm]{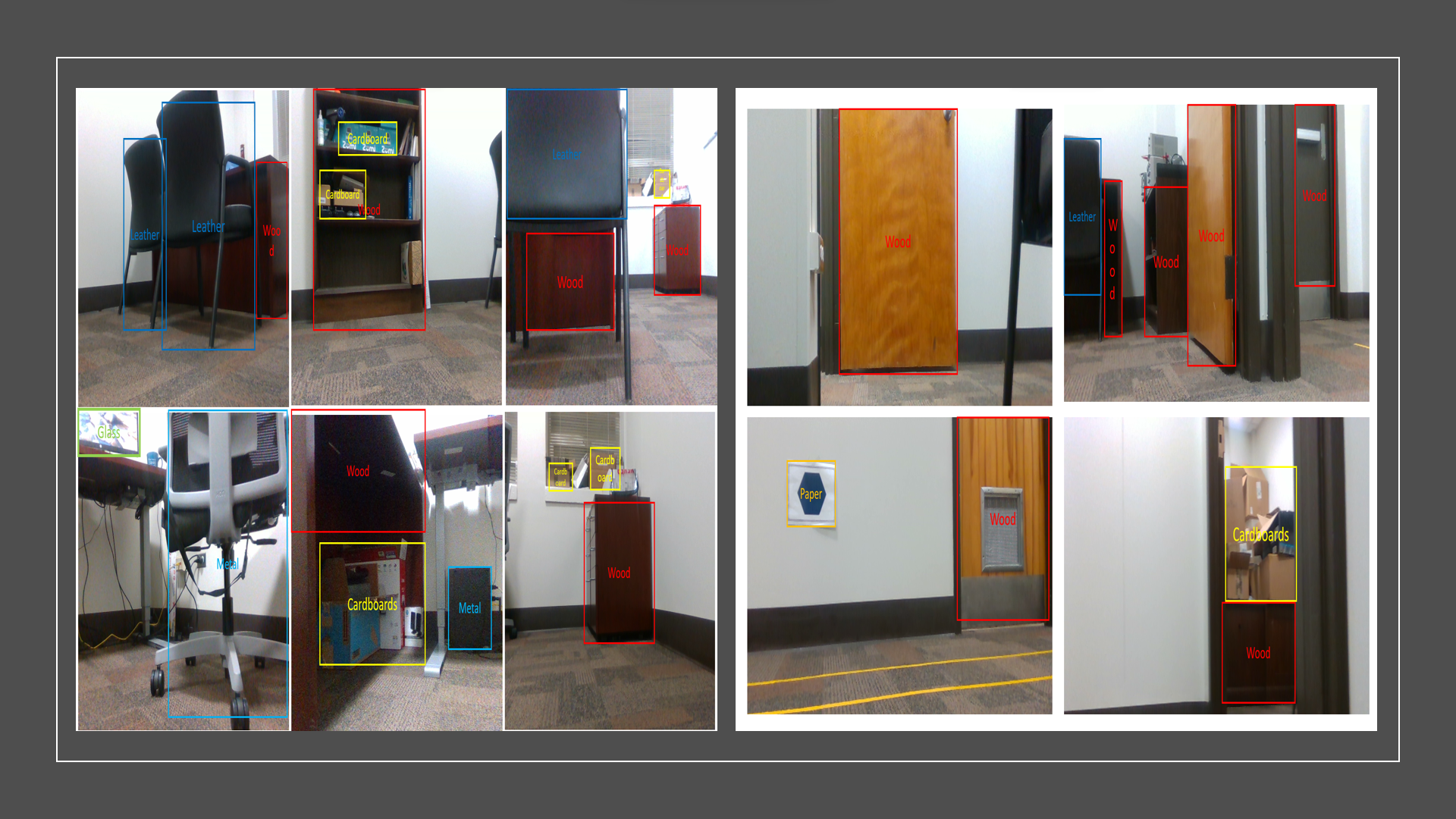}
    \caption{Office Room 1 - RGB Images, Ground Truth Materials}
    \label{fig:prof rgb}
\end{figure*}

\begin{figure*}[htp]
    \centering
    \includegraphics[width=16cm]{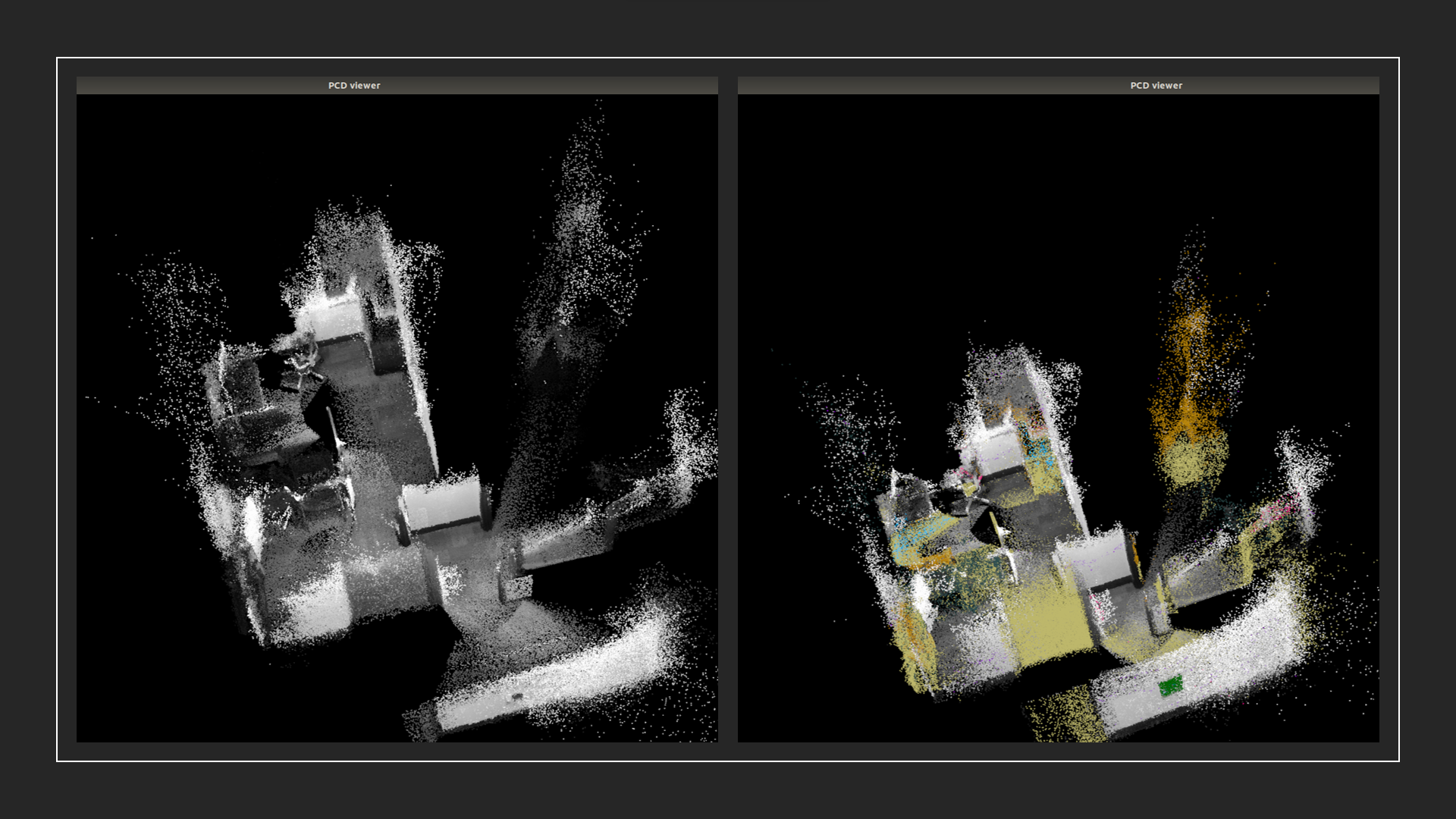}
    \caption{Office Room 1- Point Cloud and Semantic Material Map}
    \label{fig:prof room}
\end{figure*}

\begin{figure*}[htp]
    \centering
    \includegraphics[width=16cm]{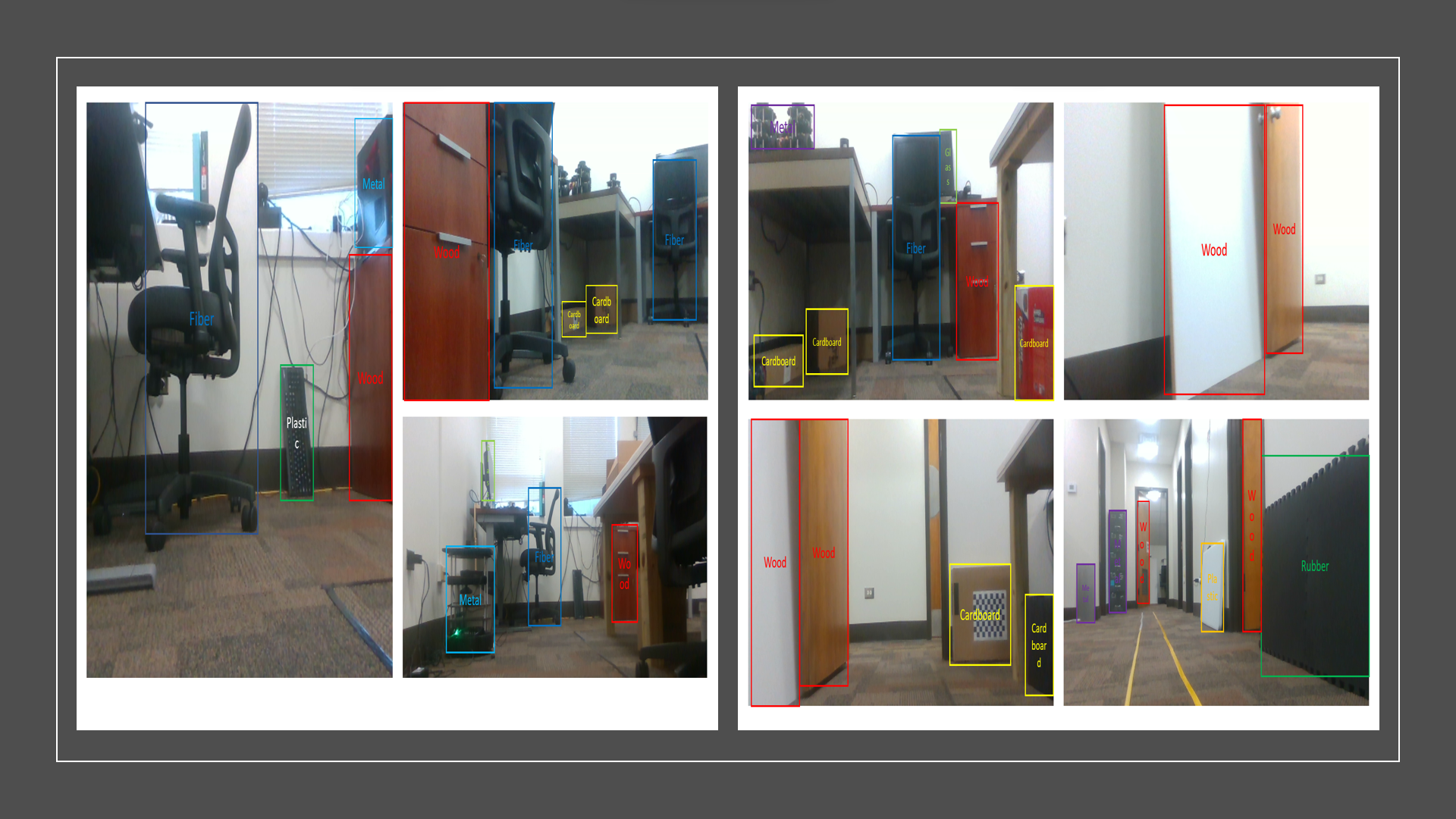}
    \caption{Laboratory Room - RGB Images and Ground Truth Materials}
    \label{fig:stat rgb}
\end{figure*}

\begin{figure*}[htp]
    \centering
    \includegraphics[width=16cm]{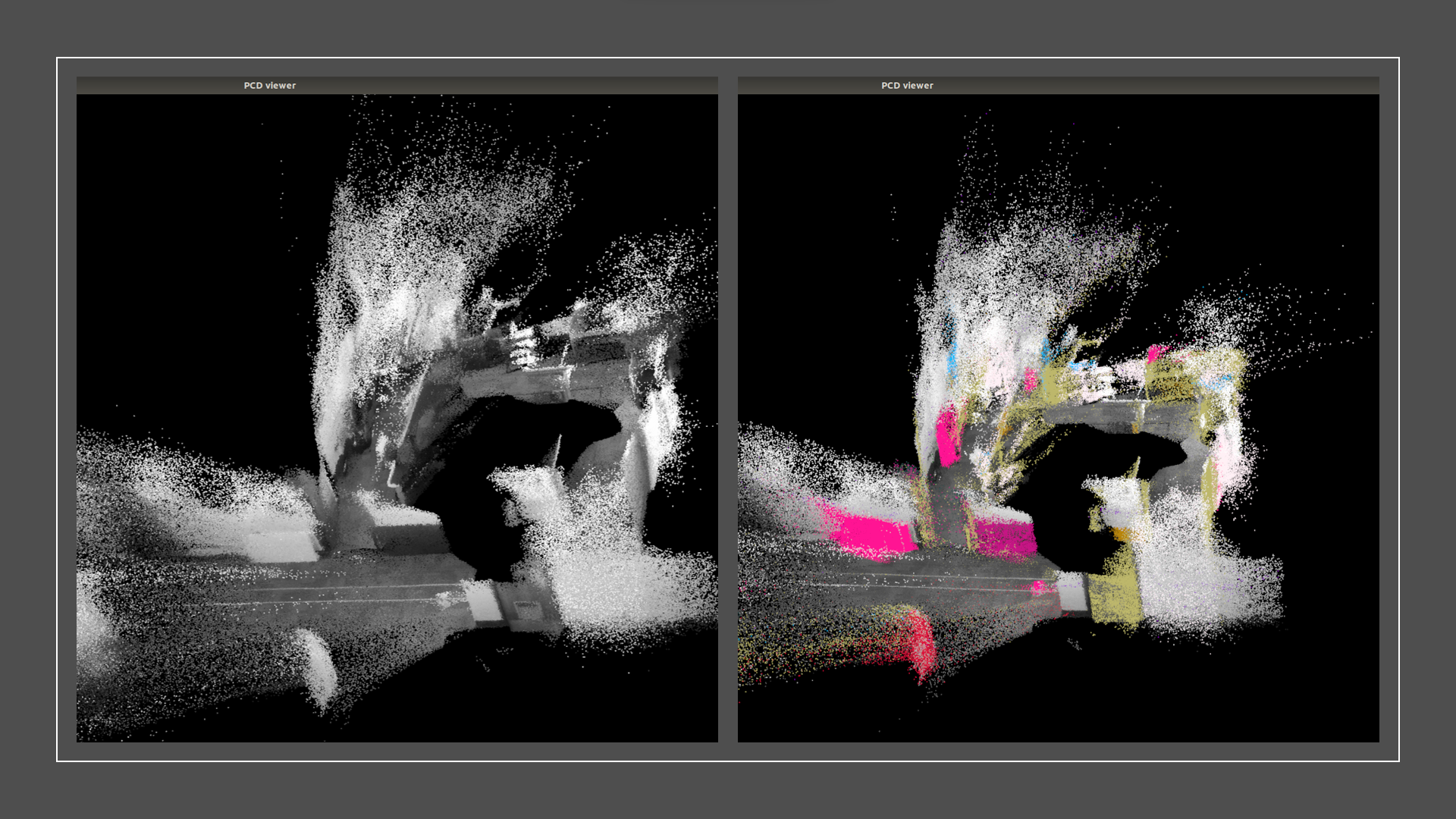}
    \caption{Laboratory Room - Point Cloud and Semantic Material Map}
    \label{fig:exp room}
\end{figure*}

\begin{figure*}[htp]
    \centering
    \includegraphics[width=16cm]{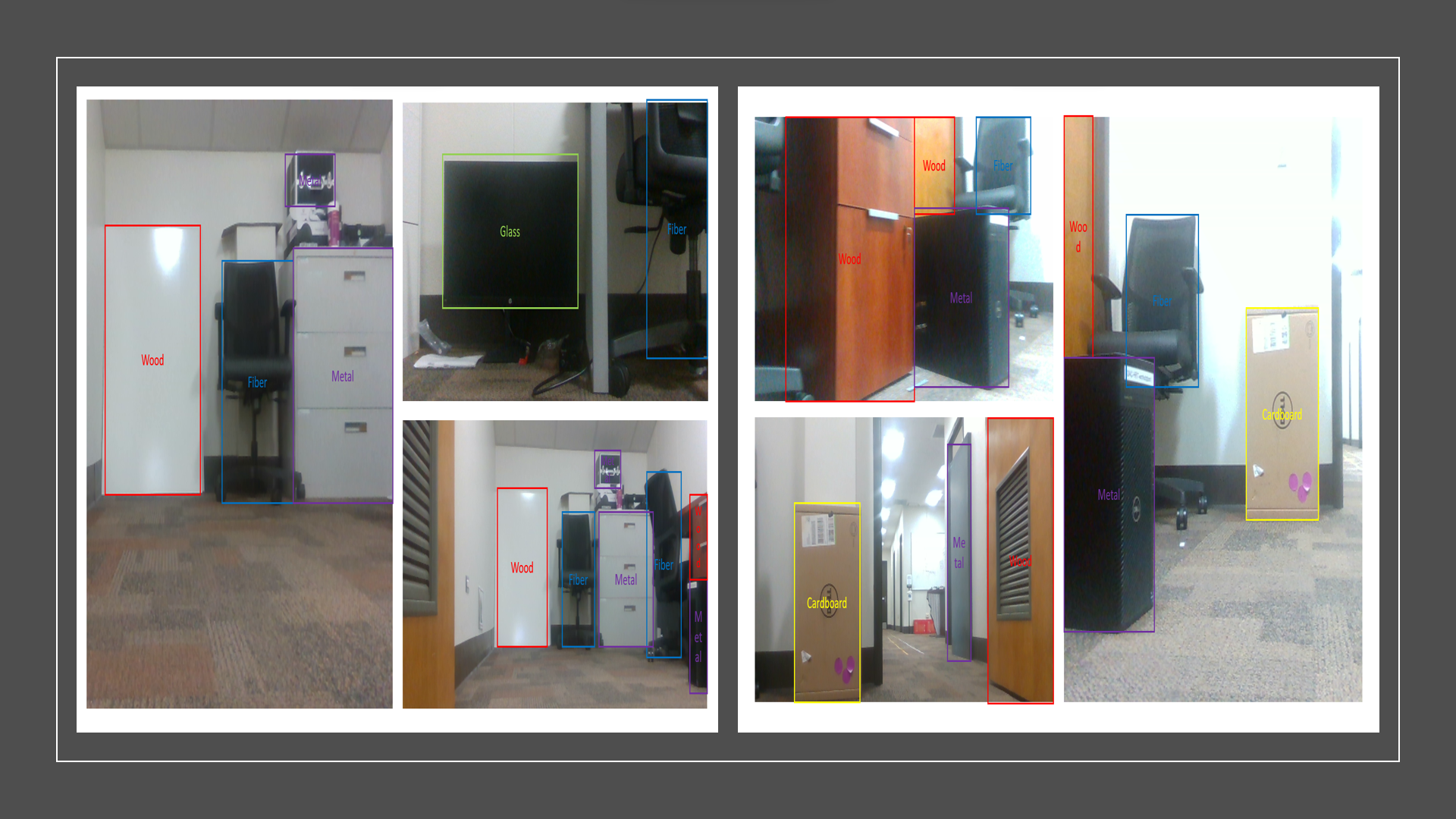}
    \caption{Office Room 2 - RGB Images and Ground Truth Materials}
    \label{fig:swarm rgb}
\end{figure*}

\begin{figure*}[htp]
    \centering
    \includegraphics[width=16cm]{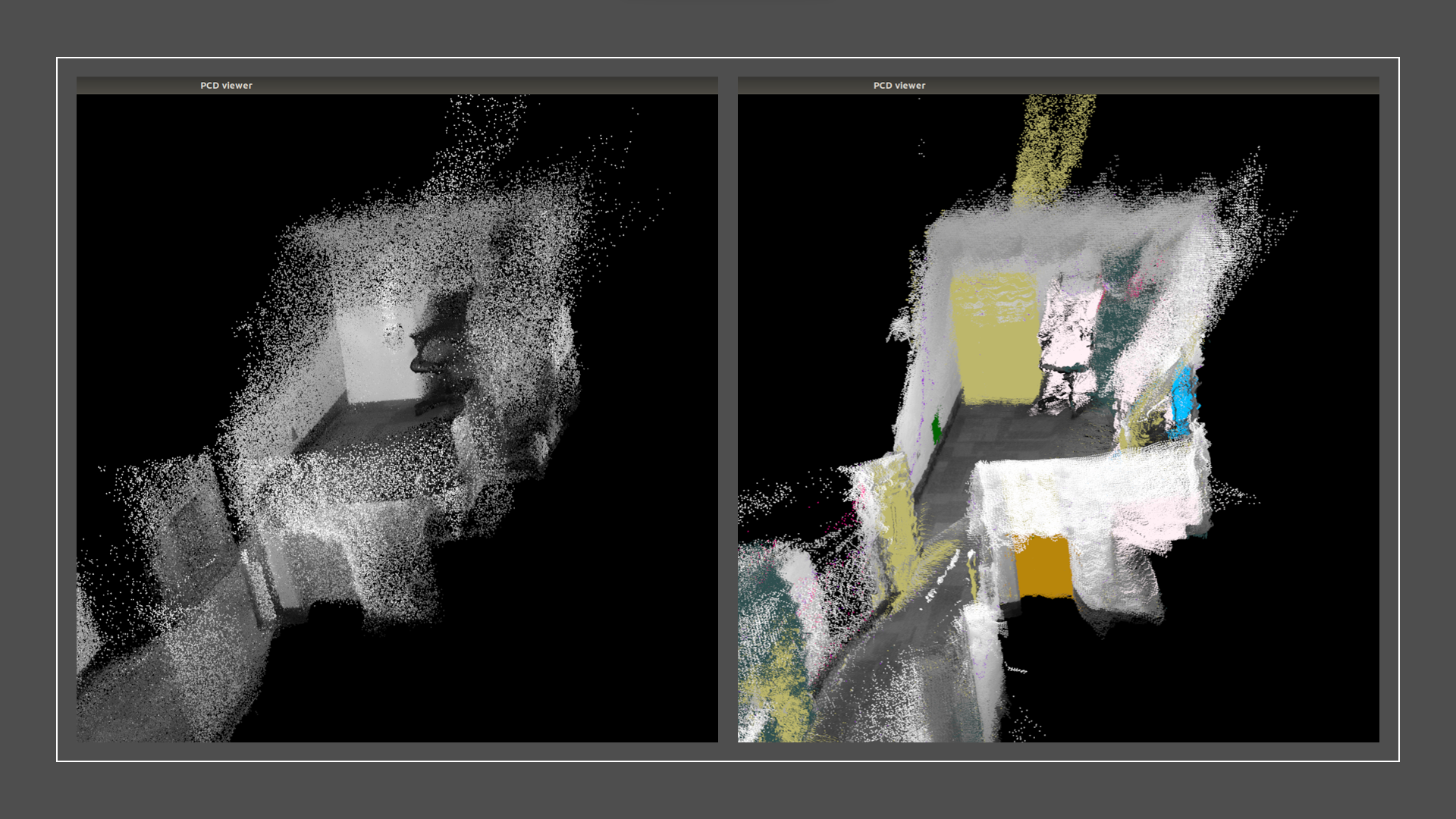}
    \caption{Office Room 2 - Point Cloud and Semantic Material Map}
    \label{fig: swarm room}
\end{figure*}

\end{document}